\documentclass{article}
\PassOptionsToPackage{numbers,compress}{natbib}

\usepackage[preprint]{neurips_2026}
\usepackage[utf8]{inputenc} 
\usepackage[T1]{fontenc}    
\usepackage{hyperref}       
\usepackage{url}            
\usepackage{booktabs}       
\usepackage{amsfonts}       
\usepackage{nicefrac}       
\usepackage{microtype}      
\usepackage{amsmath,amssymb}
\usepackage{graphicx}
\usepackage{url}
\usepackage{enumitem}
\usepackage{wrapfig}
\usepackage{float}
\usepackage{multirow}
\usepackage{makecell}
\usepackage{pifont}
\usepackage{csquotes}
\usepackage{adjustbox}
\usepackage{svg}        
\usepackage{xspace}
\usepackage{bm}
\usepackage{tabulary}
\usepackage{pgf,pgfplots}
\usepackage{tikz}
\usepackage{colortbl}
\usepackage{tcolorbox}
\usepackage{array}
\usepackage{tikz}
\usepackage{wrapfig}
\usepackage{adjustbox}
\usepackage{caption}
\usepackage{titletoc}
\usepackage{tabularx}
\usepackage{amssymb} 
\usepackage{makecell} 
\definecolor{applegreen}{rgb}{0.55, 0.71, 0.0}

\newcommand{\ccmark}{\color{applegreen} \ding{51}}%
\newcommand{\cxmark}{\color{red} \ding{55}}%

\newcommand{\benchname}{\textsc{RIG-Bench}\xspace}

\title{Thinking in Pictures: A Systematic Benchmark for Reasoning-driven Image Generation}

\author{%
  Yutong Liu$^{1,2*}$ \quad
  Nan Huang$^{1,*}$ \quad
  Xu Cao$^{1,*}$ \quad
  James M. Rehg$^{1}$ \\
  \\
  $^{1}$University of Illinois Urbana-Champaign \qquad
  $^{2}$New York University \qquad
}

\begin{document}

\maketitle

\begingroup
\renewcommand{\thefootnote}{*}
\footnotetext{Equal contribution. This work was done while Yutong Liu was an intern at UIUC.}
\endgroup

\begin{abstract}
Recent advancements in unified generative models (UGMs) and world simulators have achieved unprecedented results in visual perception and synthesis. However, these models primarily rely on surface-level event alignment, leaving the capacity for high-level visual reasoning underexplored. True visual generative intelligence demands "Reasoning-to-Generation", an ability to infer latent rules from visual inputs and manifest solutions through precise, logically constrained visual outcomes. We introduce \benchname, a novel comprehensive benchmark that systematically evaluates \textbf{R}easoning-driven \textbf{I}mage \textbf{G}eneration (\textbf{RIG}) across four cognitively demanding domains: Concept-based, Transformation-based, Pattern \& Structure, and Scenario-based. Featuring 2000 curated samples, \benchname serves as a rigorous stress test for RIG. Our extensive evaluations of state-of-the-art UGMs and image/video generation models reveal a significant reasoning-generation gap, wherein models frequently produce locally plausible but globally illogical outputs. \benchname provides a vital diagnostic framework to guide the development of next-generation, logically grounded UGMs and world simulators. \benchname is open-sourced in \href{https://huggingface.co/datasets/Abbyyyt/RIG-Bench}{huggingface.co/datasets/Abbyyyt/RIG-Bench}.
\end{abstract}

\section{Introduction}
\label{sec:introduction}

\begin{figure*}[t]
  \centering
  \includegraphics[width=\linewidth]{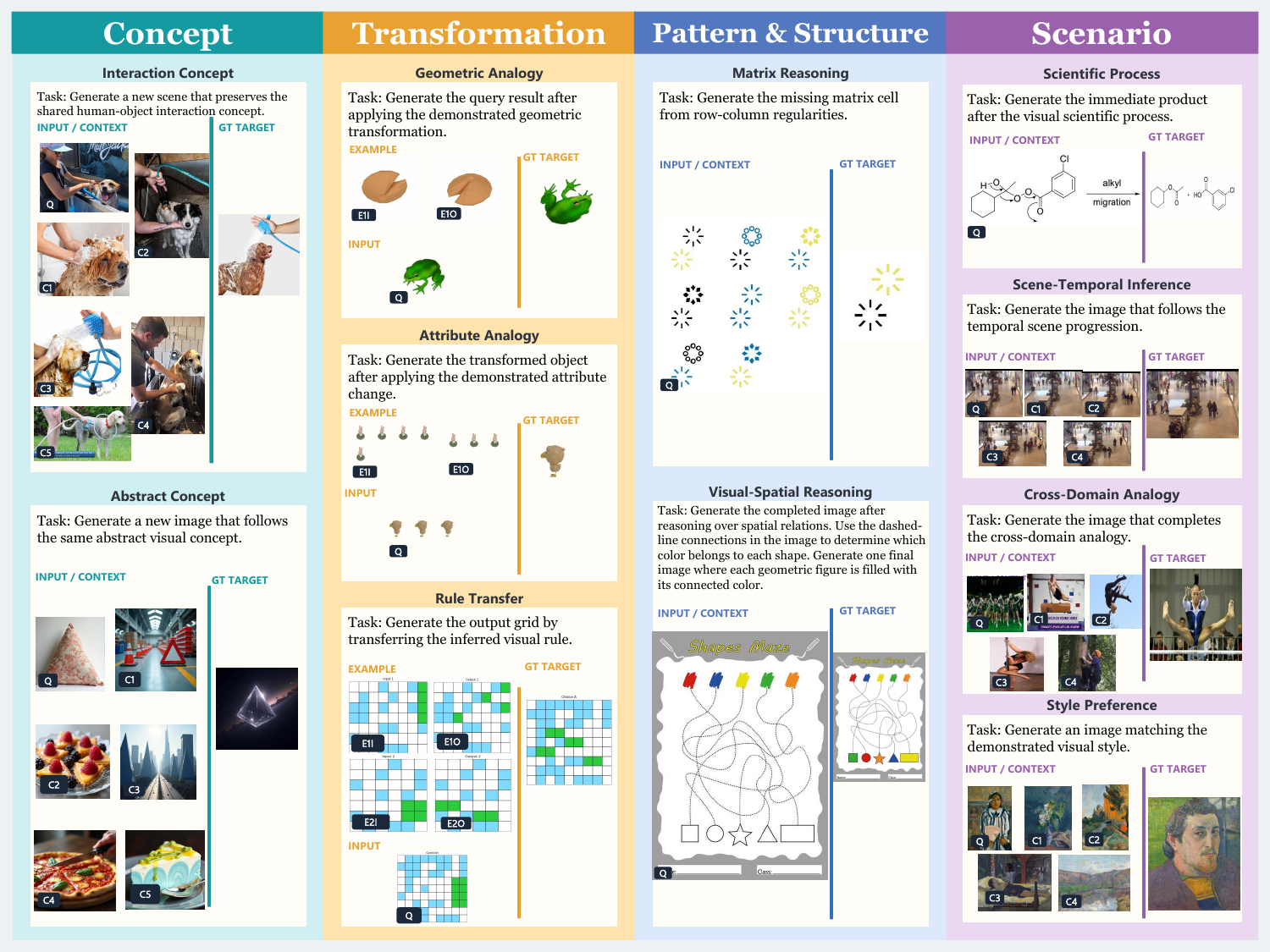}
  \caption{Overview of \benchname. Each column shows one of the four task families
  (Concept-based, Transformation-based, Pattern \& Structure, Scenario-based), illustrated with one
  representative item per fine-grained subtask. Every item supplies visual context
  (\textsc{Input/Context}) plus an unobserved ground-truth image (\textsc{GT Target});
  the model must infer the correct answer from the visual context and produce it as
  an image.}
  \label{fig:data_overview}
  \vspace{-3mm}
\end{figure*}

The recent progress in world models and unified generative models (UGMs)~\citep{unified_multimodal_survey,janus_pro,show_o,show_o2,vila_u,chameleon,emu3,bagel} has marked a paradigm shift in AI, yielding unprecedented results in visual perception and high-fidelity synthesis. Yet, these milestones primarily reflect a scaling of perceptual fluency and surface-level alignment. A more fundamental hallmark of human intelligence remains largely underexplored: the capacity for high-level reasoning over visual inputs to manifest solutions through precise visual outcomes~\citep{babyvision,vischainbench}. This ability, deeply rooted in human visual imagination and mental simulation~\citep{pearson2019human}, allows us to not only describe the world but to internalize its underlying mechanics. In both abstract and physical realms, true intelligence is defined not by recognizing a scene, but by inferring latent rules, simulating complex transformations, and predicting structured outcomes constrained by domain-specific logic. We posit that this "Reasoning-to-Generation" capability is a prerequisite for the next generation of autonomous world models.

Solving a Raven’s Progressive Matrix~\citep{raven}, navigating a complex maze~\citep{babyvision}, or predicting the product of a chemical reaction from molecular diagrams~\citep{emma} requires more than "hallucinating" a plausible image. It demands a System 2~\citep{Dual-Process} cognitive process where the model must internally simulate spatial, logical, or scientific rules. While existing models can generate visually stunning landscapes, they often falter when the output is constrained by strict logical dependencies~\citep{visulogic,vrb_primary}. For instance, a model might generate a "science-looking" diagram that violates the law of conservation of mass~\citep{mmmg,wise,phybench}, or a "maze-like" image where the path leads into a dead end. Therefore, requiring a visual answer serves as a more rigorous "stress test" for genuine multimodal reasoning than text-based VQA.

Despite its significance, this \textbf{R}easoning-driven \textbf{I}mage
\textbf{G}eneration (\textbf{RIG}) capability is not systematically
addressed by current benchmarks. Existing evaluations are bifurcated:
they either assess reasoning via text (Image-to-Text) ~\citep{mathvista,mmmu,scienceqa,visulogic} or evaluate
generation via aesthetic metrics (Text-to-Image)~\citep{geneval,t2i_compbench,dsg,vbench}, leaving a critical
gap in evaluating the closed-loop visual reasoning
(Image-to-Reasoning-to-Image) process. To bridge this gap, we introduce
\benchname, a comprehensive evaluation suite designed to probe the
high-level reasoning limits of unified multimodal models. Our benchmark
spans four cognitively demanding task families:
\emph{Concept-based} reasoning, which requires inferring a shared
concept across a set of images and generating a new instance;
\emph{Transformation-based} reasoning, which requires applying a
demonstrated geometric, attribute, or rule-level transformation to a
new input; \emph{Pattern \& Structure} reasoning, which covers matrix
completion and visual-spatial puzzles such as mazes, path tracing, and
grid-based fill-ins; and \emph{Scenario-based} reasoning, which grounds
inference in scientific processes, temporal scene progression,
cross-domain analogies, and stylistic consistency.

Through extensive experiments, we demonstrate that even the most advanced generation models (e.g., GPT Image 1, Gemini 3 Pro Image Preview, Qwen-Image) exhibit a significant reasoning–generation gap. They often produce "locally plausible but globally illogical" results, indicating that scaling alone has not yet yielded the structured reasoning required for human-like visual problem-solving.

Our contributions are summarized as follows:

\begin{itemize}[itemsep=0.5ex, parsep=0pt, topsep=-2.3pt, leftmargin=0.4cm]

\item We define the task of Reasoning-driven Image Generation (RIG), shifting the focus from aesthetic synthesis to logical-visual consistency.
\item We present \benchname, comprising $2{,}000$ curated items across $4$ task families and
$11$ fine-grained subtasks. To our knowledge, \benchname is the first benchmark in which the target output is never specified in language and must be induced from visual context alone, across concept, transformation, pattern, and scenario domains under a single answer-image protocol.
\item We provide a diagnostic evaluation framework that reveals the failure modes of current UGMs and image/video generation models, offering insights into the future design of "thinking" generative agents.

\end{itemize}

\section{Related Works}

\paragraph{Visual Generation.}
Visual generation has advanced toward increasingly high-quality and
controllable synthesis~\citep{esser2024scaling}, while evaluation has
expanded beyond perceptual quality to prompt faithfulness,
compositionality, and temporal consistency~\citep{geneval,t2i_compbench,
t2i_compbench_plus,dsg,vbench,vbench++}. Most existing benchmarks,
however, still treat generation as prompt-conditioned rendering.
Reasoning-aware T2I benchmarks~\citep{r2i_bench,mmmg,beyond_words_pixels,
ureason,mind_brush} introduce commonsense, knowledge, or other forms of
reasoning, but formulate the reasoning problem primarily in language:
the generator receives either explicit reasoning or a prompt from which
the required reasoning can be derived. Reasoning-oriented image editing
benchmarks~\citep{RISEbench,gir_bench} additionally provide visual input,
but pair it with an explicit editing instruction that specifies the
desired transformation. In both settings, what should be generated is
largely specified through language. In contrast, \benchname withholds
the target transformation and requires unified generation models to
infer what should be generated from visual context before
rendering the answer.

\paragraph{Datasets for Visual Reasoning and Cognition.}
Visual reasoning benchmarks have long studied the inference problems
\benchname targets, including abstract rule induction in ARC~\citep{arc}, matrix reasoning in Raven-style tasks~\citep{raven}, visual analogies and
concept learning~\citep{kiva,bongard_hoi,bongard_rwr}, and scientific or
multidisciplinary reasoning~\citep{scienceqa,mathvista,mmmu,emma,
mars_vqa,mmr_life}. More recent cognitive benchmarks probe mental
visualization and foundational visual gaps in MLLMs, including
Hyperphantasia~\citep{hyperphantasia}, MentisOculi~\citep{mentisoculi},
and VisFactor~\citep{visfactor}. These works expose important limits in
visual cognition, but most rely on text,
multiple-choice, or classification outputs, which verify conceptual
selection but cannot test whether a model can synthesize the inferred
answer. \benchname reframes visual reasoning as answer-image generation,
requiring the model to visually synthesize the inferred answer rather
than merely identify it.

\paragraph{Unified Multimodal Generation Models.}
The rise of unified models combining understanding and
generation~\citep{janus_pro,bagel,emu3,show_o,show_o2,vila_u} makes it
practical to bridge reading visual context and synthesizing visual
answers. Recent work has begun coupling reasoning with
generation~\citep{unigrpo,uni_mmmu,mme_unify,gap_eval}, treating
generation as a substrate for broader knowledge
tasks~\citep{babyvision,v_reasonbench,vigor_bench,beyond_words_pixels},
suggesting that benchmarks must move beyond image quality to measure
whether generation is grounded in correct structural inference.
\benchname advances this through a unified answer-image generation
protocol: models must parse visual context, infer latent rules, and
synthesize the answer image.

\section{Benchmark Design}
\label{sec:dataset}

\begin{figure*}[t]
  \centering
  \includegraphics[width=\linewidth]{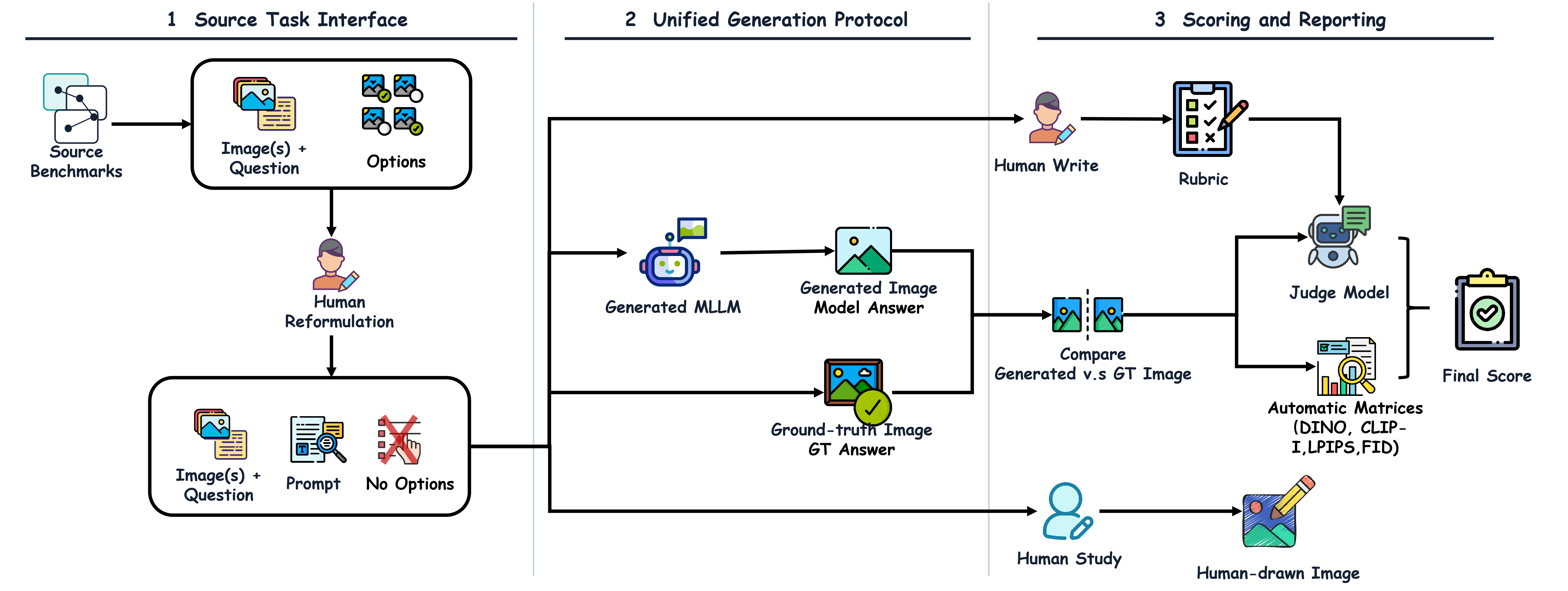}
  \caption{End-to-end pipeline of \benchname. \textbf{(1) Source Task Interface:} items are drawn from established visual reasoning benchmarks and manually re-cast with an image-based output prompt (no options exposed to the model). \textbf{(2) Unified Generation Protocol:} selected models receive the context images plus the prompt and generate the answer image directly. \textbf{(3) Scoring and Reporting:} each generation is compared against the ground-truth image using both an LLM judge over a hand-written rubric and automatic perceptual metrics, with an additional human study for calibration.}
  \label{fig:pipeline}
  \vspace{-3mm}
\end{figure*}


\subsection{Task Formalization}
\label{sec:task_def}

We redefine visual generation as a problem of \textit{latent rule induction} rather than mere instruction following. In this framework, \emph{Reasoning-driven Image Generation} (RIG) is formulated as a unified synthesis task. A model is provided with a visual context $\mathcal{C} = (\mathcal{I}, \mathcal{D}, t)$, where:
\begin{enumerate}[label=(\roman*), leftmargin=*]
    \item $\mathcal{I} = \{i_1, \ldots, i_k\}$ is a set of $k \geq 1$ context images that implicitly encode a visual reasoning problem (e.g., a pattern or a spatial configuration);
    \item $\mathcal{D} = \{(a_j, b_j)\}_{j=1}^{m}$ is optional. It consists of zero or more ordered demonstration pairs exemplifying a target transformation rule (active primarily in transformation-based tasks);
    \item $t$ is a natural-language instruction that specifies the output constraints (e.g., ``complete the sequence'') without disclosing the underlying logic or the step-by-step solution.
\end{enumerate}

The goal is to synthesize a single output image $\hat{y} \in \mathbb{R}^{H \times W \times 3}$ that constitutes the logically correct solution. Correctness is evaluated by comparing $\hat{y}$ to a ground-truth image $y$ across two dimensions: \textbf{perceptual fidelity} (visual quality) and \textbf{logical consistency} (adherence to the induced rule).

This unified formulation mandates three essential, inter-dependent competencies that are often decoupled in existing benchmarks:
\begin{itemize}[leftmargin=*]
    \item \textbf{Visual Perception:} The model must perform fine-grained feature extraction—identifying attributes such as shape, color, cardinality, and spatial topology—from the context $\mathcal{I}$.
    \item \textbf{Rule Induction:} The model must infer the abstract relationship or transformation latent within $\mathcal{I}$ or demonstrated by $\mathcal{D}$. This requires moving beyond surface-level pattern matching to high-level logical abstraction.
    \item \textbf{Grounded Generation:} The model must manifest its internal reasoning by rendering the solution into pixels. Unlike text-based VQA, this stage requires the model to maintain structural and logical integrity during the high-dimensional synthesis process.
\end{itemize}


\begin{wrapfigure}{r}{0.40\textwidth}
    \centering
    \vspace{-2em}
    \includegraphics[width=0.98\linewidth]{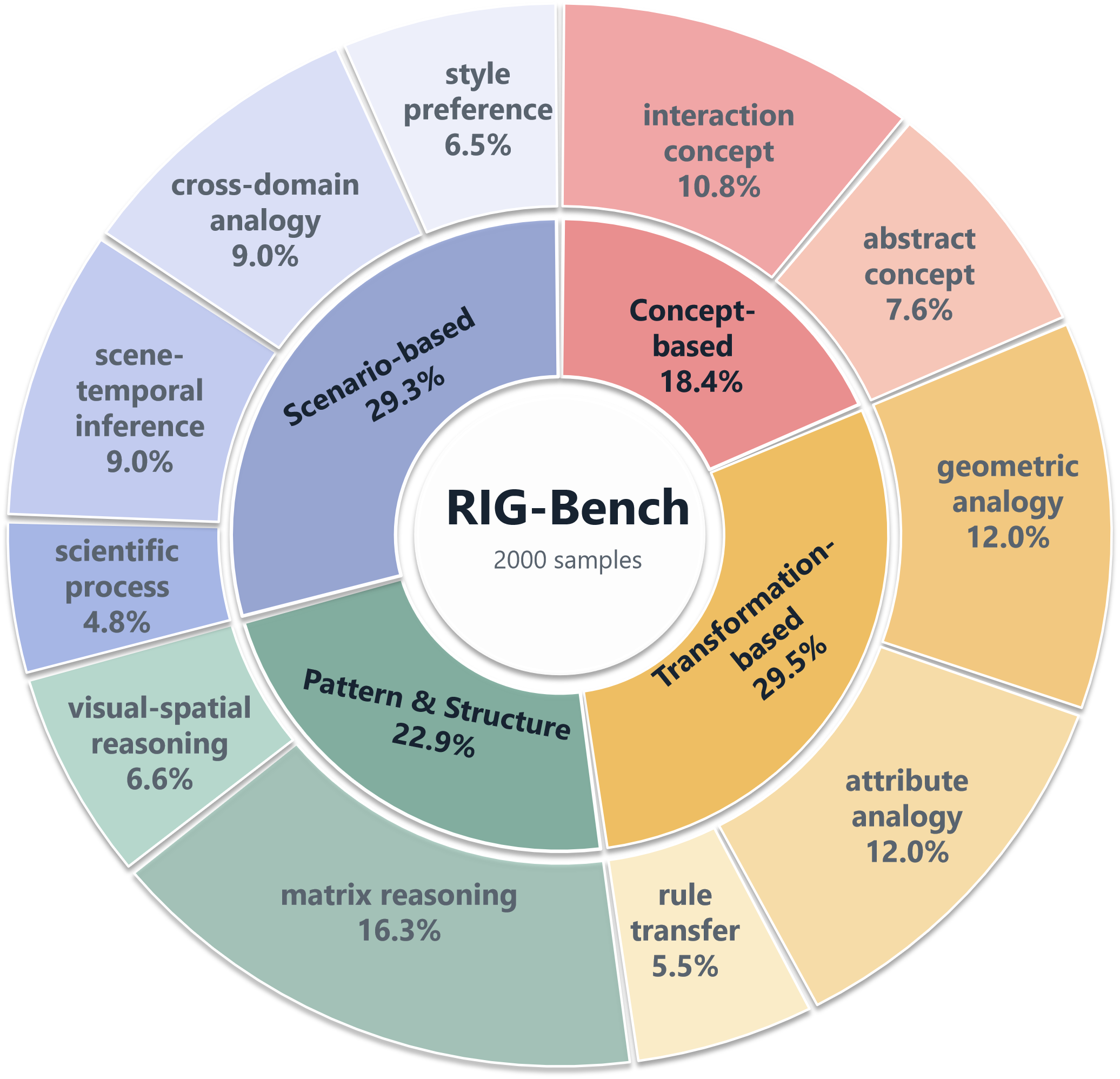}
    \label{fig:subtask_distribution}
    \caption{Distribution of the 2,000 samples in \benchname. The inner ring summarizes the four task families, and the outer ring breaks them down into eleven fine-grained subtasks.}
    \vspace{-3em}
\end{wrapfigure}

\begin{figure*}[t]
  \centering
  \includegraphics[width=\linewidth]{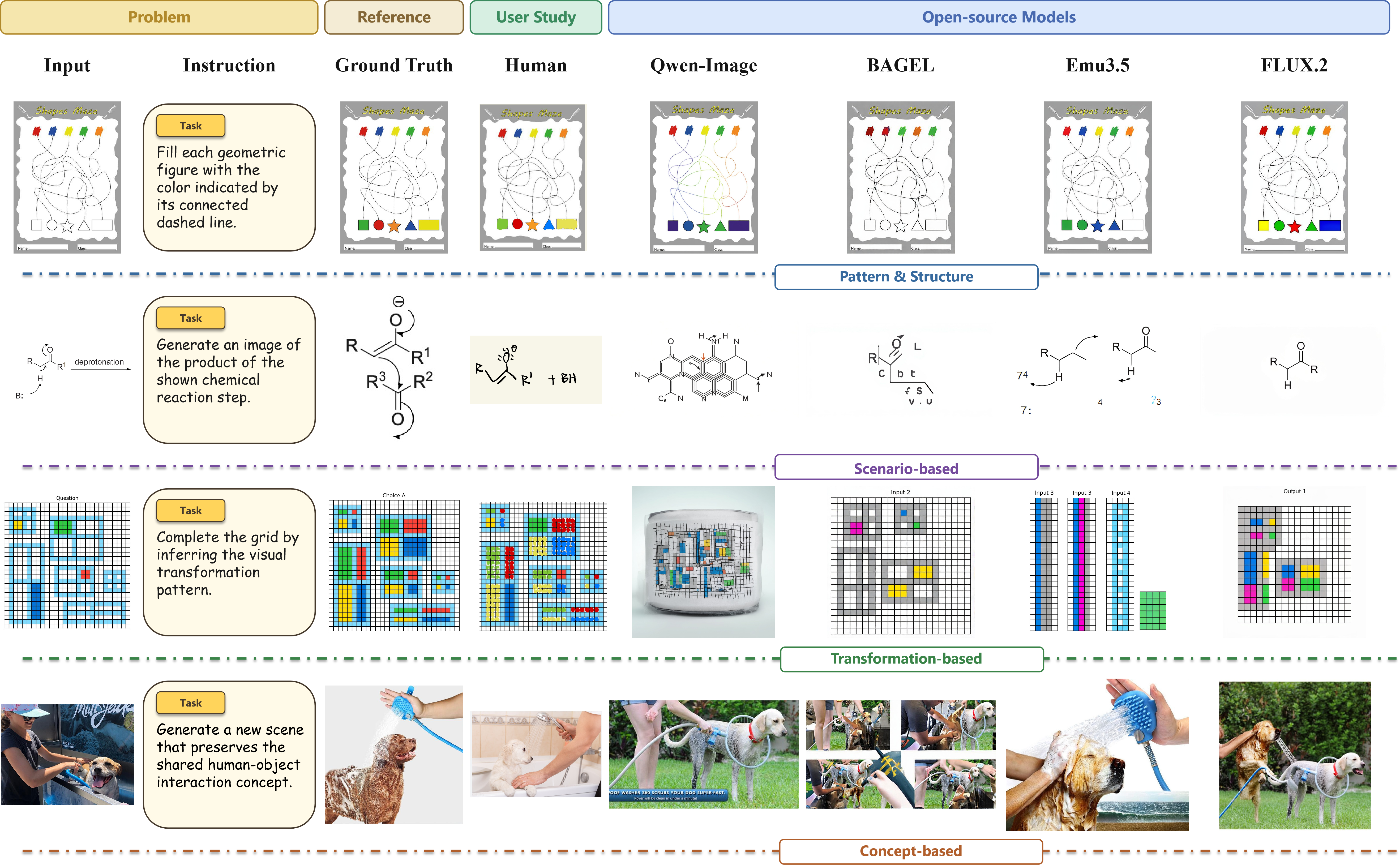}
  \includegraphics[width=\linewidth]{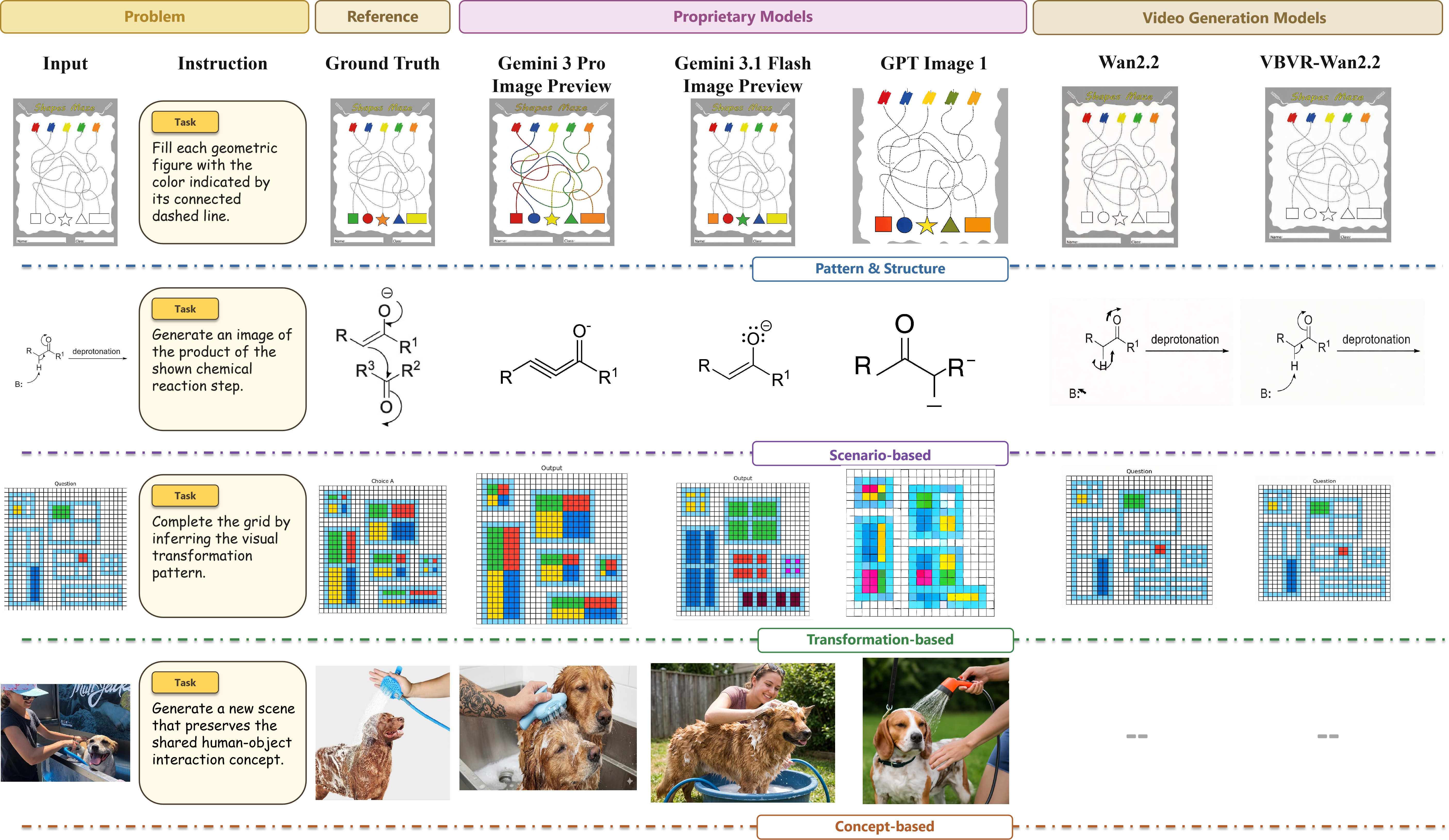}
  \label{fig:qualitative_open}
  \vspace{-1em}
\caption{Example across the four task families of \benchname, one
representative subtask per row.
Model outputs grouped into open-source
(top) and proprietary plus video generators (bottom). 
}
\vspace{-1.5em}
\end{figure*}
While previous benchmarks evaluate perception and induction through multiple-choice or text-based answers, \benchname is the first to require all three competencies simultaneously. By using the synthesized image as the primary evidence of reasoning, we provide a more rigorous diagnostic for the ``reasoning--generation gap'' in world models and unified generative models~\citep{ureason,gir_bench}.


\subsection{Data Engineering and Curation}
\label{sec:data}

\noindent\textbf{High-Fidelity Item Distillation.}
Unlike text-based reasoning tasks, pixel-level reasoning demands an uncompromising approach to curation to prevent shortcut learning. We employ a three-tier distillation process:
\emph{(i)~Cross-Domain Reformulation.} We distill core reasoning logic
from diverse sources~\citep{bongard_hoi, kiva, arc, emma, bongard_rwr, mars_vqa, mmr_life, babyvision} and
re-conceptualize them as generative tasks, shifting the objective from
recognition to synthesis and requiring models to manifest latent rules
(e.g., concept induction, structural analogy, or spatial inference)
through precise visual outputs. Sourcing from established datasets
inherits prior calibration on what constitutes a meaningful inference
problem, while shifting the discriminating axis from inter-option
confusability to a joint function of rule induction and rendering
precision.
\emph{(ii)~Reasoning-Neutral Prompting.} We manually author instructions $t$ for all samples. To ensure the benchmark evaluates pure reasoning rather than instruction following, these prompts define the output constraints (e.g., ``generate the missing matrix cell'') without providing any semantic cues or logical shortcuts that might simplify the rule-induction process.
\emph{(iii)~Heuristic and Human Filtering.} The initial pool is pruned to eliminate low-complexity or perception-only samples. We enforce a ``human-solvable but model-challenging'' criterion, with item difficulty informed by source-dataset design and checked through a formative human pilot.
This pipeline ensures that \benchname serves as a diagnostic stress test for the reasoning--generation gap. 
By consolidating validated challenges from multiple domains, \benchname achieves greater complexity and logical rigor than its constituent sources.

\noindent\textbf{Dataset Statistics and Structural Variation.}
\benchname comprises \textbf{2,000} high-quality samples, characterized by a deliberate variability in input cardinality to suit diverse reasoning motifs. As illustrated in Figure~\ref{fig:data_overview}, the number of context images is systematically scaled by task family:
\textit{(i)~Concept-based} items provide six instances that collectively define an abstract category; 
\textit{(ii)~Transformation-based} items supply between one and six demonstration pairs $\{\mathcal{D}\}$ alongside the query image; 
\textit{(iii)~Pattern \& Structure} tasks utilize a single composite puzzle (e.g., a $3{\times}3$ Raven's-style matrix); 
and \textit{(iv)~Scenario-based} items include one to five real-world images depicting a complex state. 
Crucially, despite this input diversity, the output is strictly constrained to a single synthesized image $\hat{y}$. This design ensures a \textbf{uniform evaluation interface} across all subtasks. To prevent model anchoring or leakage of the reasoning chain, all instructions $t$ are authored as neutral, declarative directives (e.g., \textit{``Generate the image that completes the pattern.''}), ensuring the model relies solely on its internal induction capabilities.

\noindent\textbf{Taxonomic Diversity and Skill Coverage.}
Table~\ref{tab:subtask_distribution} reports the distribution across
the four families and eleven subtasks. \textbf{Transformation-based}
and \textbf{Scenario-based} form the core of the benchmark:
Transformation tasks evaluate rule induction through geometric
analogies (rotation, reflection), attribute shifts (color, scale), and
pixel-grid rule transfer; Scenario tasks pair perceptual grounding with
world knowledge, covering temporal inference, cross-domain analogy, and
scientific structural synthesis. \textbf{Pattern \& Structure} targets
systematic pattern induction via $3{\times}3$ matrix reasoning and
visual-spatial panels, while \textbf{Concept-based} probes few-shot
abstract grouping.

\begin{wraptable}{r}{0.48\columnwidth}
    \centering
    \vspace{-4mm}
    \captionof{table}{Sub-task distribution of \benchname across four task families and eleven
    fine-grained subtasks. Reasoning skill coverage spans concept induction,
    visual transformations, pattern reasoning, and scenario-based inference.}
    \label{tab:subtask_distribution}
    \scriptsize
    \setlength{\tabcolsep}{4pt}
    \renewcommand{\arraystretch}{0.95}
    \resizebox{0.95\linewidth}{!}{%
    \begin{tabular}{ll c c}
    \toprule
    \textbf{Task Family} & \textbf{Sub-task} & \textbf{\# Samples} & \textbf{Ratio (\%)} \\
    \midrule
    \multirow{2}{*}{Concept-based}
      & interaction concept       & 216 & 10.8 \\
      & abstract concept          & 152 & 7.6  \\
    \midrule
    \multirow{3}{*}{Transformation-based}
      & geometric analogy         & 240 & 12.0 \\
      & attribute analogy         & 240 & 12.0 \\
      & rule transfer             & 109 & 5.5  \\
    \midrule
    \multirow{2}{*}{Pattern \& Structure}
      & matrix reasoning          & 326 & 16.3 \\
      & visual-spatial reasoning  & 131 & 6.6  \\
    \midrule
    \multirow{4}{*}{Scenario-based}
      & scientific process        & 96  & 4.8  \\
      & scene-temporal inference  & 180 & 9.0  \\
      & cross-domain analogy      & 180 & 9.0  \\
      & style preference          & 130 & 6.5  \\
    \midrule
    \textbf{Total} & -- & \textbf{2000} & \textbf{100.0} \\
    \bottomrule
    \end{tabular}
    }
    \vspace{-4mm}
\end{wraptable}

\noindent\textbf{Answer-Space Constraint.}
Beyond the four-family taxonomy shown in Figure~\ref{fig:data_overview},
\benchname also varies whether the target image is open-form or
closed-form. Open-form subtasks allow multiple visually valid realizations
of an inferred concept, style, or scene continuation, as in interaction
concept, abstract concept, scene-temporal inference, and style preference.
Closed-form subtasks require a rule-constrained target determined by a
geometric, attribute, structural, scientific, or analogy relation, covering
geometric analogy, attribute analogy, rule transfer, matrix reasoning,
visual-spatial reasoning, scientific process, and cross-domain analogy.
This distinction lets us separate failures on permissive generative
continuation from failures on tightly constrained visual reasoning.

\subsection{Comparison with Existing Benchmarks}
\label{sec:comparison}

A robust world model must achieve a closed-loop transition from visual perception to visual prediction through three synergistic functions: \textbf{(i) Perceptual Grounding}: perceiving a complex scene; \textbf{(ii) Latent Rule Induction}: inferring the underlying logic linking inputs to outputs; and \textbf{(iii) Visual Synthesis}: manifesting the predicted state in high-fidelity pixels. We categorize existing multimodal benchmarks based on these three functional axes (Table~\ref{tab:benchmark_comparison}), revealing a critical gap in the current evaluation landscape\citep{mme_unify,uni_mmmu,gap_eval}.

\begin{table*}[h]
\centering
\caption{Comparison of \benchname against existing benchmarks. A
visual reasoning loop requires perceiving a complex scene
(\textbf{Visual Context}), inferring the underlying logic
(\textbf{Rule Induction}), and manifesting the solution in pixels
(\textbf{Visual Synthesis}). Existing benchmarks address only a subset.}
\vspace{-0.5em}
\label{tab:benchmark_comparison}
\scriptsize
\begin{tabularx}{\textwidth}{l c c c p{4.5cm} l}
\toprule
\textbf{Benchmark Camp} & \makecell{\textbf{Visual}\\\textbf{Context}} & \makecell{\textbf{Rule}\\\textbf{Induction}} & \makecell{\textbf{Visual}\\\textbf{Synthesis}} & \textbf{Skill Scope} & \makecell{\textbf{Representative}\\\textbf{Works}} \\
\midrule
Text-to-image (T2I) & \cxmark & \cxmark & \ccmark & Compositional \& Aesthetic & \cite{geneval,t2i_compbench} \\
\addlinespace[0.5em]
Reasoning-aware T2I & \cxmark & \ccmark & \ccmark & Semantic constraints & \cite{commonsense_t2i,t2i_reasonbench} \\
\addlinespace[0.5em]
VQA / Visual Reasoning & \ccmark & \ccmark & \cxmark & Analytical \& Cognitive & \cite{mmmu,mathvista,arc,bongard_hoi} \\
\addlinespace[0.5em]
Image Editing & \ccmark & \cxmark & \ccmark & Local attribute shifts & \cite{instructpix2pix,magicbrush} \\
\addlinespace[0.5em]
\midrule
\rowcolor{gray!5}
\textbf{\benchname (Ours)} & \ccmark & \textbf{\ccmark} & \textbf{\ccmark} & \textbf{Cognitive, Scientific, \& World Knowledge} & --- \\
\bottomrule
\end{tabularx}
\end{table*}

Standard \textit{Text-to-Image (T2I) Generation}~\citep{geneval, t2i_compbench, dsg} focuses exclusively on synthesis, lacking both visual contextual input and reasoning-driven constraints. While \textit{Reasoning-aware T2I}~\citep{commonsense_t2i, t2i_reasonbench} introduces logical steps, these are typically mediated through text, bypassing the challenges of visual-spatial perception. Conversely, \textit{VQA and Visual Reasoning} benchmarks~\citep{mmmu, mathvista, arc, bongard_hoi} evaluate complex induction and perception but collapse the output space into low-dimensional text or categorical labels, failing to test the model's ability to "think in pictures." 


\textit{Image Editing} benchmarks~\citep{instructpix2pix, magicbrush} occupy the visual-in/visual-out space but generally rely on explicit local instructions rather than the discovery of latent rules from context. The most related efforts involve \textit{Visual-input Video Reasoning}~\citep{v_reasonbench, vbvr, vigor_bench}; however, these are predominantly limited to procedurally synthesized environments and video-specific architectures. \benchname uniquely demands the simultaneous execution of all three functions: perception, induction, and synthesis, across a diverse array of hand-curated, cognitively demanding tasks. It serves as a unified diagnostic for world models and UGMs, pushing beyond surface-level alignment toward true visual intelligence.

\section{Experiments and Results}
\label{sec:experiments}

\subsection{Experimental Setup}

\noindent \textbf{Models.} We evaluate recent multimodal generation models on all eleven subtasks of
\benchname. For open-source models, we consider
Qwen-Image~\citep{qwen_image}, BAGEL~\citep{bagel}, Emu3.5~\citep{emu3}, and FLUX.2~\citep{flux2_dev}. For proprietary
image-generation models, we evaluate Gemini~3 Pro Image Preview, Gemini
3.1 Flash Image Preview~\citep{gemini31_pro,gemini31_flash_image} and 
GPT~Image~1~\citep{gpt_image_1}. Although \benchname defines the
target as an answer image, video-generation models are also relevant because their final frames can be interpreted as predicted visual states after reasoning over the input context. We therefore evaluate  Wan2.2~\citep{wan2.2}, and the reasoning-augmented variant
VBVR-Wan2.2~\citep{vbvr}.

\begin{table*}[t]
\caption{Main results on \benchname. Bold indicates the best result within each model group. Common Rubric is computed on the nine subtasks supported by all evaluated model families and reports 95\% bootstrap confidence intervals. Full Rubric is computed on each model's full supported evaluation set.}
\vspace{-0.5em}
\label{tab:main_results}
\centering
\scriptsize
\setlength{\tabcolsep}{4pt}
\resizebox{\textwidth}{!}{%
\begin{tabular}{lcccccc}
\toprule
\textbf{Models} & \makecell{\textbf{Common}\\\textbf{Rubric} $\uparrow$} & \makecell{\textbf{Full}\\\textbf{Rubric} $\uparrow$} & \textbf{DINO} $\uparrow$ & \textbf{CLIP} $\uparrow$ & \textbf{LPIPS} $\downarrow$ & \textbf{FID} $\downarrow$\\
\midrule
\multicolumn{7}{c}{\textbf{Image Generation Models} (Open-source)} \\
Qwen-Image~\citep{qwen_image} & 20.79 {\scriptsize [19.91, 21.69]} & 25.71 & 0.39 & 0.70 & 0.66 & 77.38 \\
BAGEL~\citep{bagel} & 16.76 {\scriptsize [16.29, 17.25]} & 16.68 & 0.44 & 0.70 & 0.67 & 97.46 \\
Emu3.5~\citep{emu3} & 20.66 {\scriptsize [19.91, 21.39]} & 21.48 & \textbf{0.46} & \textbf{0.72} & \textbf{0.61} & \textbf{62.85} \\
FLUX.2~\citep{flux2_dev} & \textbf{26.35 {\scriptsize [25.57, 27.14]}} & \textbf{31.31} & 0.43 & \textbf{0.72} & \textbf{0.61} & 87.29 \\
\midrule
\multicolumn{7}{c}{\textbf{Image Generation Models} (Proprietary)} \\
Gemini 3 Pro Image Preview~\citep{gemini31_pro} & \textbf{59.71 {\scriptsize [58.65, 60.77]}} & \textbf{64.57} & \textbf{0.58} & \textbf{0.79} & \textbf{0.57} & \textbf{63.25} \\
Gemini 3.1 Flash Image Preview~\citep{gemini31_flash_image} & 53.40 {\scriptsize [51.99, 54.80]} & 58.08 & 0.52 & 0.76 & 0.61 & 75.79 \\
GPT Image 1~\citep{gpt_image_1} & 37.57 {\scriptsize [36.34, 38.85]} & 43.77 & 0.49 & 0.75 & 0.60 & 66.44 \\
\midrule
\multicolumn{7}{c}{\textbf{Video Generation Models}} \\
Wan2.2-14B~\citep{wan2.2} & 20.73 {\scriptsize [19.99, 21.48]} & 20.73 & 0.58 & 0.80 & 0.50 & 72.06 \\
VBVR-Wan2.2-14B~\citep{vbvr} & \textbf{21.56 {\scriptsize [20.90, 22.23]}} & \textbf{21.56} & \textbf{0.63} & \textbf{0.82} & \textbf{0.46} & \textbf{65.84} \\
\bottomrule
\end{tabular}
}
\end{table*}

\begin{table*}[t]
\caption{Sub-task-level rubric scores. Bold indicates the best per-group result for each subtask.}
\vspace{-0.5em}
\label{tab:subtask_llm_judge}
\centering
\scriptsize
\setlength{\tabcolsep}{3.5pt}
\renewcommand{\arraystretch}{0.95}
\resizebox{\textwidth}{!}{%
\begin{tabular}{lccccccccccc}
\toprule
\multirow{2}{*}{\textbf{Models}}
& \multicolumn{2}{c}{\textbf{Concept-based}}
& \multicolumn{3}{c}{\textbf{Transformation-based}}
& \multicolumn{2}{c}{\textbf{Pattern \& Structure}}
& \multicolumn{4}{c}{\textbf{Scenario-based}} \\
\cmidrule(lr){2-3}
\cmidrule(lr){4-6}
\cmidrule(lr){7-8}
\cmidrule(lr){9-12}
& \textbf{Inter.}
& \textbf{Abs.}
& \textbf{Geo.}
& \textbf{Attr.}
& \textbf{Rule}
& \textbf{Matrix}
& \textbf{V-Space}
& \textbf{Sci.}
& \textbf{Scene}
& \textbf{Cross}
& \textbf{Style} \\
\midrule
\multicolumn{12}{c}{\textbf{Image Generation Models} (Open-source)} \\
Qwen-Image & 49.88 & \textbf{44.23} & 13.82 & 11.29 & 14.20 & 18.33 & 29.82 & 8.72 & 28.06 & \textbf{26.81} & 44.44 \\
BAGEL      & 17.94 & 14.08 & 11.09 & \textbf{16.55} & 17.39 & 15.84 & 27.40 & 3.50 & 15.93 & 9.82 & 39.24 \\
Emu3.5     & 16.75 & 36.65 & 10.44 & 8.80 & 10.53 & 18.00 & \textbf{39.40} & 10.74 & \textbf{38.78} & 17.95 & 44.38 \\
FLUX.2     & \textbf{64.72} & 37.60 & \textbf{17.08} & 10.98 & \textbf{24.27} & \textbf{34.33} & 32.37 & \textbf{14.77} & 37.45 & 11.53 & \textbf{61.90} \\
\midrule
\multicolumn{12}{c}{\textbf{Image Generation Models} (Proprietary)} \\
Gemini 3 Pro Image Preview     & \textbf{92.08} & \textbf{75.88} & \textbf{52.54} & \textbf{69.94} & 52.57 & \textbf{57.63} & \textbf{68.51} & \textbf{65.55} & 34.28 & \textbf{62.43} & 90.47 \\
Gemini 3.1 Flash Image Preview & 85.98 & 69.45 & 50.29 & 64.12 & \textbf{58.94} & 36.65 & 60.55 & 41.19 & \textbf{58.72} & 54.43 & \textbf{91.57} \\
GPT Image 1                    & 76.78 & 63.55 & 29.59 & 22.34 & 40.69 & 36.93 & 27.73 & 25.05 & 31.31 & 50.85 & 90.51 \\
\midrule
\multicolumn{12}{c}{\textbf{Video Generation Models}} \\
Wan2.2-14B      & N/A & N/A & 24.67 & 10.87 & 22.89 & 16.80 & 27.04 & 11.48 & 25.93 & \textbf{20.41} & \textbf{33.37} \\
VBVR-Wan2.2-14B & N/A & N/A & \textbf{25.85} & \textbf{12.59} & \textbf{25.81} & \textbf{17.85} & \textbf{33.17} & \textbf{12.98} & \textbf{26.83} & 20.18 & 25.15 \\
\bottomrule
\end{tabular}
}
\vspace{0.5pt}
\begin{flushleft}
\scriptsize
\textit{Abbreviations:} Inter. = interaction concept; Abs. = abstract concept; Geo. = geometric analogy; Attr. = attribute analogy; Rule = rule transfer; Matrix = matrix reasoning; V-Space = visual-spatial reasoning; Sci. = scientific process; Scene = scene-temporal inference; Cross = cross-domain analogy; Style = style preference. ``N/A'' indicates unsupported settings: Concept-based subtasks require
multi-image inputs, whereas the evaluated video-generation models support only
single-image conditioning.
\end{flushleft}
\vspace{-2.5em}
\end{table*}

\noindent \textbf{Implementation Details.} All models are queried under the unified protocol of
Section~\ref{sec:task_def}: the model receives the input images and a
natural-language instruction and must produce a single answer image.
Source multiple-choice options are withheld so the model must generate
the answer rather than select it. We use one generation per sample at
each model's default sampling settings, and apply no post-processing.
For video-generation models, we instruct the model to treat the generated clip as a reasoning trajectory: the preceding frames may unfold the inference process visually, and the {final frame} is designated as the answer image used for evaluation. 
For open-source models, experiments are conducted on NVIDIA H200 GPUs
(141\,GB HBM3e).

\subsection{Evaluation Metrics}
\label{sec:metrics}

\noindent \textbf{Reference-based perceptual metrics.}
We report sample-level DINO~\citep{dinov2} and CLIP-I~\citep{clip}
cosine similarities (DINOv2-base CLS, CLIP ViT-B/32; higher is better)
and LPIPS~\citep{lpips} perceptual distance (AlexNet; lower is better)
between each generation and its ground-truth answer, plus a
distribution-level FID~\citep{fid} computed over the 2,000 samples. These quantify visual proximity but are
agnostic to whether the generation reflects correct reasoning.

\noindent \textbf{LLM-as-a-judge with Pre-defined Rubric.}
For every subtask we hand-craft a 0--5 rubric on three dimensions:
\emph{visual quality}, \emph{structural alignment}, and \emph{reasoning
correctness}. The first two share common anchors across all subtasks;
reasoning-correctness anchors are subtask-specific. Following the rubric-based evaluation protocol validated in UEval~\citep{ueval,healthbench},
we employ Gemini-3.1-Flash as the judge and query it three times per sample at temperature 0.9 and average dimension scores. The composite score is
$\mathrm{Score} = 0.15\cdot\mathrm{VQ} + 0.20\cdot\mathrm{SA} +
0.65\cdot\mathrm{RC}$, with reasoning correctness dominant because it
is the capability the benchmark targets. Composites are rescaled to
$[0,100]$ for reporting. All rubrics are released with the benchmark.

\noindent \textbf{Validation of judge reliability.}
To validate the reliability of our automatic evaluation protocol, we conduct a blinded human study on 500 benchmark items stratified across all 11 subtasks, with twelve annotators evaluating outputs from all nine models using the same rubric dimensions as the automatic judge. Human ratings show substantial inter-annotator agreement on the composite score (Krippendorff's $\alpha=0.791$). The main Gemini-3.1-Flash judge shows strong agreement with human ratings (Pearson/Spearman $=0.766/0.711$), while an independent non-Gemini Qwen3.5-27B judge achieves comparable agreement ($0.791/0.700$). The two automatic judges also agree strongly with each other ($0.819/0.791$). Model rankings are also largely preserved under the non-Gemini judge
(Kendall $\tau=0.833$ with the Gemini judge). Consistent results (Table ~\ref{tab:judge_human_agreement}) across ICC, Cohen's $\kappa$, and model-ranking agreement further support the reliability of the automatic evaluation protocol; detailed statistics are reported in Appendix~\ref{app:human_validation}.

\begin{table}[h]
\centering
\caption{Agreement among automatic judges and blinded human ratings. Gemini
denotes the main Gemini-3.1-Flash judge; Qwen3.5 denotes the Qwen3.5-27B judge
used for cross-checking.}
\label{tab:judge_human_agreement}
\small
\setlength{\tabcolsep}{6pt}
\begin{tabular}{lccc}
\toprule
Metric & Gemini vs. Human & Qwen3.5 vs. Human & Gemini vs. Qwen3.5 \\
\midrule
Pearson correlation & 0.766 & 0.791 & 0.819 \\
Spearman correlation & 0.711 & 0.700 & 0.791 \\
ICC(2,1) & 0.764 & 0.789 & 0.817 \\
Cohen's $\kappa$ & 0.634 & 0.644 & 0.650 \\
Model ranking (Kendall $\tau$) & 1.000 & 0.833 & 0.833 \\
\bottomrule
\end{tabular}
\end{table}

\subsection{Main Results}
\label{sec:main_results}

\begin{wrapfigure}{h}{0.45\textwidth}
    \centering
    \includegraphics[width=\linewidth]{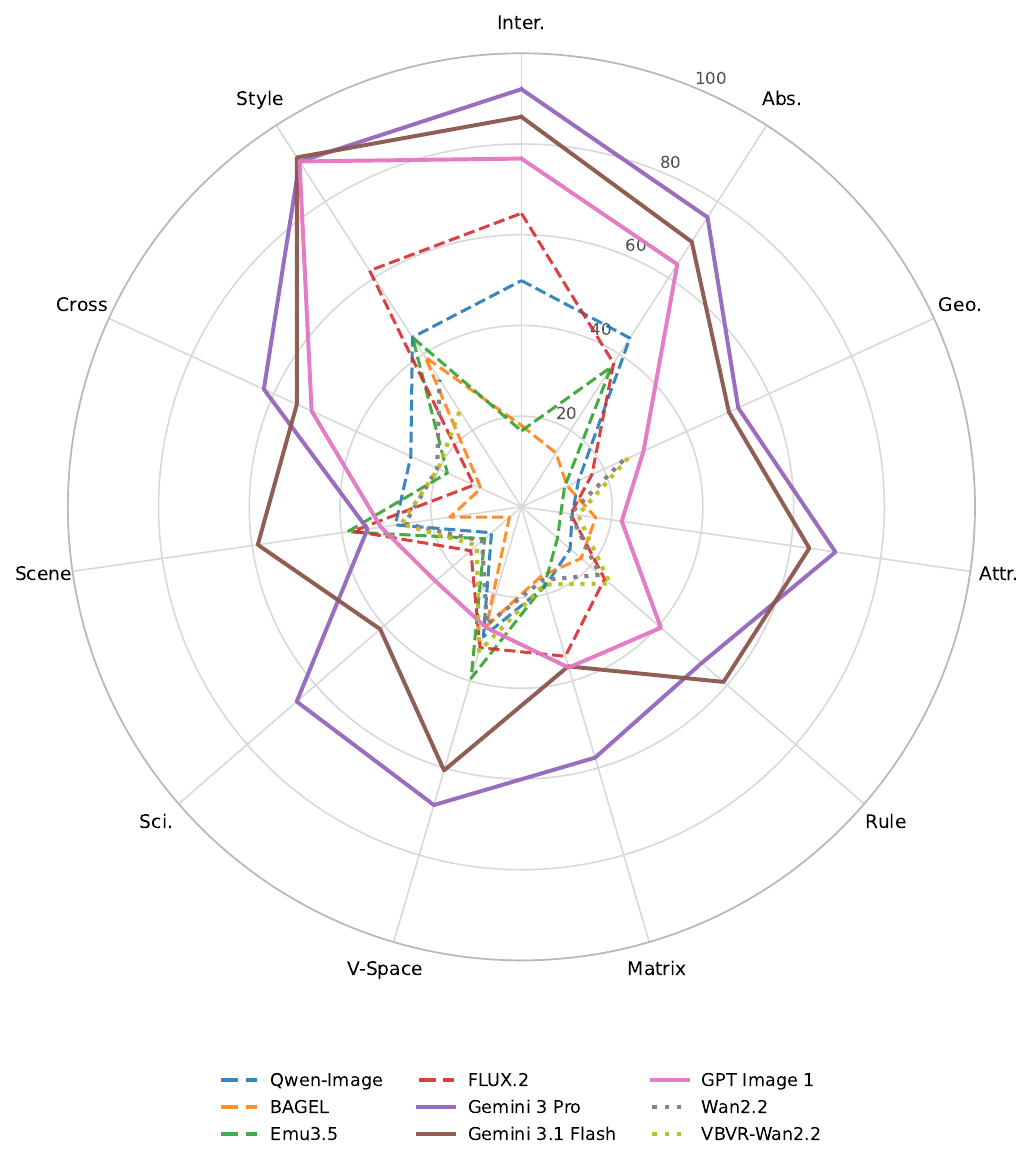}
    \caption{\textbf{Per-subtask rubric scores across nine generation
    models on \benchname.} Line style indicates model group: solid for
    proprietary image generators, dashed for open-source image
    generators, dotted for video generators. }
    \label{fig:subtask_radar}
\end{wrapfigure}

Table~\ref{tab:main_results} and Figure~\ref{fig:subtask_radar} report common-supported and full-supported rubric scores together with reference-based metrics; Table~\ref{tab:subtask_llm_judge}
disaggregates the rubric by subtask.

\noindent \textbf{No model approaches saturation, while humans solve
the tasks reliably.}
The strongest model, Gemini~3~Pro Image Preview, attains a composite of
\textbf{64.6/100}; every other proprietary model sits in the $40$--$60$
range, and every open-source image generator scores below $32$. No model consistently exceeds $70$ across the benchmark. 

Separately, before finalizing the benchmark, we conduct a formative
human pilot on $n{=}66$ items uniformly sampled across the eleven
subtasks, in which participants infer the correct answer and either
draw it or retrieve a matching image from the web. Human outputs are
manually checked for answer correctness, with the LLM judge used as an
auxiliary rubric-based assessment. Participants achieve {92.9\%}
manual accuracy and an average LLM-judge score of {97.3}. We use this
pilot as a descriptive solvability check rather than a population-level
human ceiling, indicating that the tasks are reliably solvable when the
underlying visual reasoning is performed correctly. Notably, several source datasets
of \benchname are nearing ceiling under their original closed-form
protocols~\citep{kiva, bongard_hoi}: replacing multiple-choice with
image-output reopens the difficulty gap.

\begin{table}[h]
\centering
\caption{Rubric scores grouped by answer-space constraint on the
common-supported subtasks. Open-form includes scene-temporal inference and
style preference; closed-form includes the seven rule-constrained subtasks.
Concept-based subtasks are omitted here to match the common-supported scope
shared by image and video generators.}
\label{tab:answer_form_split}
\small
\setlength{\tabcolsep}{5pt}
\resizebox{0.75\linewidth}{!}{%
\begin{tabular}{lccc}
\toprule
Model & Open-form & Closed-form & Gap \\
\midrule
\multicolumn{4}{c}{\textbf{Image Generation Models} (Open-source)} \\
Qwen-Image & 34.9 & 17.5 & +17.4 \\
BAGEL & 25.7 & 14.7 & +11.0 \\
Emu3.5 & 41.1 & 15.9 & +25.2 \\
FLUX.2 & \textbf{47.7} & \textbf{21.4} & +26.3 \\
\midrule
\multicolumn{4}{c}{\textbf{Image Generation Models} (Proprietary)} \\
Gemini 3 Pro Image Preview & 57.8 & \textbf{60.8} & -3.0 \\
Gemini 3.1 Flash Image Preview & \textbf{72.5} & 51.1 & +21.4 \\
GPT Image 1 & 56.1 & 33.4 & +22.8 \\
\midrule
\multicolumn{4}{c}{\textbf{Video Generation Models}} \\
Wan2.2-14B & \textbf{29.1} & 18.8 & +10.3 \\
VBVR-Wan2.2-14B & 26.1 & \textbf{20.5} & +5.6 \\
\bottomrule
\end{tabular}
}
\end{table}

\noindent \textbf{Performance stratifies sharply across models and
subtasks.}
The strongest open-source generator (FLUX.2, $31.3$) trails the weakest
proprietary one (GPT~Image~1, $43.8$) by $12.5$ points and is
$33.3$ points below Gemini~3~Pro ($64.6$). This open--proprietary gap
is substantially larger than what the same models exhibit on
text-conditioned generation benchmarks~\citep{geneval, t2i_compbench_plus},
suggesting that the visual-input--visual-output regime exposes capability
differences that surface alignment to text prompts otherwise masks.
Stratification is equally sharp across subtasks: Concept-based and
\emph{style preference} items often score several times higher than
the hardest rule-induction settings for open-source systems, with
Gemini~3~Pro reaching $92.1$ on \emph{interaction concept} and above
$90$ on \emph{style preference}, while the strongest open-source scores
remain below $25$ on the three Transformation-based subtasks. Tasks with
open-ended answer spaces are forgiving; tasks whose answer is uniquely
determined by a latent rule are punishing. The Transformation family is
the discriminating spine of \benchname, the family on which genuine
rule-induction capacity, rather than open-ended generative plausibility,
separates models.


\noindent \textbf{Perceptual similarity does not track reasoning
correctness.}
A central motivation for \benchname is that perceptual similarity metrics
conflate visual proximity with reasoning correctness, and our results
quantify this dissociation directly~\citep{gap_eval,ureason,gir_bench}. All nine evaluated models achieve
non-trivial similarity to the ground-truth answer in perceptual feature
space (DINO $\geq 0.39$, CLIP-I $\geq 0.70$), yet their rubric composites
span a $3.9{\times}$ range, from $16.7$ for BAGEL to $64.6$ for
Gemini~3~Pro. VBVR-Wan2.2 is the extreme case within the video group: it
attains strong DINO ($0.63$), CLIP-I ($0.82$), LPIPS ($0.46$), and FID
($65.8$), yet a Rubric of $21.6$, below most image-generation baselines.
Video clips, by virtue of temporal consistency,
often produce frames pixel-wise close to the ground truth even when the
underlying reasoning is wrong; a leaderboard based on DINO/CLIP alone
would conclude that video generators are state-of-the-art reasoners,
which they manifestly are not.
Section~\ref{sec:analysis} dissects this gap with matched text- and
image-output conditions.

\subsection{Analysis: Where does end-to-end generation break?}
\label{sec:analysis}
Aggregate scores show that models fail on \benchname, but not whether
the failure arises from visual inference, visual realization, or their
coupling. We therefore run a matched diagnostic analysis on the same
200 stratified items, covering all four families and all 11 subtasks,
under four conditions: \emph{T}, a text or closed-form answer inferred
from the original visual context; \emph{D}, direct answer-image
generation from the original visual context, which is the paper's
protocol; \emph{H}, image generation conditioned on the model's own
text answer; and \emph{O}, image generation conditioned on an oracle
answer specification. D/H/O are judged blindly against the ground-truth
image with the image rubric. T uses a separate text-correctness rubric,
so it diagnoses text-side reasoning rather than serving as a direct
image-quality score.

\begin{table}[h]
\centering
\caption{Matched diagnostic decomposition across text-reasoning and
image-generation conditions on 200 stratified items. Scores are reported
in the same 0--100 unit as Table~\ref{tab:main_results}. Brackets show
paired-bootstrap 95\% CIs for O--D. Integration gap is the fraction of
items with T and O passing but D failing at $\tau{=}3$.}
\vspace{0.5em}
\label{tab:matched_decomposition}
\small
\setlength{\tabcolsep}{4pt}
\resizebox{\textwidth}{!}{%
\begin{tabular}{lcccccc}
\toprule
Model setting & T & D & H & O & O--D & Integration gap \\
\midrule
BAGEL (unified) & 41.3 & 33.6 & 39.7 & 46.0 & +12.4 [+9.4, +15.5] & 5.5\% \\
Qwen3.5 $\rightarrow$ Qwen-Image-Edit & 49.9 & 37.3 & 36.4 & 39.0 & +1.8 [-1.8, +5.4] & 4.5\% \\
Qwen3.5 $\rightarrow$ Emu3.5-Image & 50.8 & 32.8 & 46.2 & 52.0 & +19.3 [+15.5, +23.2] & 9.5\% \\
Gemini~3.1~Flash Image Preview & 52.7 & 59.1 & 66.5 & 64.0 & +4.9 [-0.1, +9.9] & 9.9\% \\
\bottomrule
\end{tabular}
}
\vspace{-0.5em}
\end{table}

\noindent \textbf{Explicit reasoning helps some systems, but not uniformly.}
Conditioning image generation on the model's own inferred answer (H)
improves over direct generation (D) for BAGEL (+6.1), the
Qwen3.5$\rightarrow$Emu3.5-Image pipeline (+13.4), and Gemini
(+7.4), but not for Qwen3.5$\rightarrow$Qwen-Image-Edit (-0.9).
Thus, making the inferred answer explicit can alleviate some
end-to-end failures, but the benefit depends strongly on the generator.

\noindent \textbf{Oracle conditioning reveals rendering and integration gaps.}
The O--D column asks whether performance improves when the answer is
specified explicitly. The effect is heterogeneous. BAGEL improves by
12.4 points and the Emu3.5 pipeline by 19.3 points, while the
Qwen-Image-Edit pipeline changes by only 1.8 points with a CI crossing
zero. Because the two Qwen pipelines share the same Qwen3.5 reasoner and
the same textual plans, this contrast isolates the renderer: a correct
or useful plan only helps if the image generator can consume and realize
it faithfully.

\noindent \textbf{End-to-end evaluation exposes an integration gap.}
We define an integration-gap item as one where the system reasons
correctly in text and renders correctly when given the verified answer,
yet fails in direct image generation (T passes, O passes, D fails).
Such cases occur on 4.5--9.9\% of matched items across systems. A
pipeline of existing image-to-reasoning and oracle-conditioned
reasoning-to-image tests would certify these cases as solved, while
\benchname marks the actual visual output as wrong. This is the specific failure mode targeted by reasoning-driven image
generation: discovering what should be generated from visual evidence
and realizing it in pixels must succeed within the same end-to-end task,
without access to the answer specification. Additional intervention and per-subtask
diagnostics are provided in Appendix~\ref{app:diagnostic}.


\section{Conclusion}

We introduced \benchname, a large-scale systematic benchmark designed to evaluate Reasoning-driven Image Generation (RIG) with 2,000 samples. By shifting the evaluation paradigm from "Instruction Following" to "Rule Induction," \benchname provides a rigorous diagnostic for a model's ability to "think in pictures." Our findings expose a critical weakness in current UGMs: the inability to bridge the gap between high-level logical inference and low-level visual synthesis. We hope \benchname will serve as a foundational tool for the community, guiding the development of the next generation of world models that are not only visually stunning but logically consistent.




\bibliographystyle{plainnat}
\bibliography{main}

\clearpage
\newpage
\appendix
\clearpage

\section{Generation Prompts}
\label{app:prompts}

This appendix lists all prompts used at inference time. Each call to the model
under test combines a fixed \emph{general inference template}
(Section~\ref{app:template}) with one \emph{task-specific instruction}
(Section~\ref{app:task_prompts}) selected by the sample's subtask.

\subsection{General Inference Template}
\label{app:template}

For every sample, the API content list is assembled in the order shown below.
Demonstration pairs are present only for transformation-based subtasks; the
optional task-description suffix is included only when the source item carries
a natural-language hint.

\begin{tcolorbox}[colback=gray!4, colframe=gray!50, boxrule=0.4pt, arc=1mm, left=4pt, right=4pt, top=3pt, bottom=3pt]
\small\ttfamily
Below are $N$ demonstration input-output pair(s) showing the transformation rule:\\
\hspace*{1em}--- Example 1 input ---\\
\hspace*{1em}\textnormal{$\langle$demonstration input image(s)$\rangle$}\\
\hspace*{1em}--- Example 1 output ---\\
\hspace*{1em}\textnormal{$\langle$demonstration output image(s)$\rangle$}\\
\hspace*{1em}\textnormal{\textit{$\ldots$ additional example pairs as needed}}\\
=== End of examples. The following is the TEST INPUT. Apply the same rule and generate the corresponding output image. ===\\[3pt]
\textnormal{$\langle$test input image(s)$\rangle$}\\[3pt]
\textnormal{$\langle$\textsc{task-specific instruction}~(see Section~\ref{app:task_prompts})$\rangle$}\\[3pt]
Task description:\\
\hspace*{1em}\textnormal{$\langle$task-description text$\rangle$}
\end{tcolorbox}

The model is queried with \texttt{response\_modalities=["TEXT", "IMAGE"]} and
must return at least one image part; multiple-choice options, when present in
the source item, are intentionally withheld so the model must synthesise the
answer rather than select it. Each model is run once per sample at its
default sampling configuration; for video-generation models we take the final
frame of the returned clip as the answer image.

\subsection{Task-Specific Instructions}
\label{app:task_prompts}

We list the verbatim instructions slotted into the template above, organised
by task family and subtask. Within a subtask, multiple instruction variants
are used when the underlying activity admits distinct natural framings; every
variant follows the same general template.

\paragraph{Concept-based / Interaction concept.}
\begin{quote}\itshape\small
You are given a set of images that all depict the same abstract visual
concept --- a specific type of interaction, activity, or relationship present
in every scene. Study all provided images and identify the shared abstract
concept. Then generate a new image that clearly depicts the same concept in a
visually distinct but conceptually consistent scenario. Output a single image
that unambiguously belongs to the same category as all the provided examples.
\end{quote}

\paragraph{Concept-based / Abstract concept.}
\begin{quote}\itshape\small
You are given a set of real-world images that all share an underlying
abstract concept or relational pattern. Each image captures a different scene
or object, yet all exemplify the same higher-level idea. Study all provided
images to identify the shared concept, then generate a new image that clearly
displays that same concept in a different but valid context. Output a single
image that unambiguously belongs to the same category as all the examples.
\end{quote}

\paragraph{Transformation-based / Geometric analogy, Attribute analogy, and Rule transfer (single-pair items).}
A single shared instruction is used for the geometric and attribute analogy
subtasks, and for items in the rule transfer subtask that present a single
demonstration pair followed by a query:
\begin{quote}\itshape\small
You are solving a visual analogy puzzle. You will first see a demonstration
pair: a source object and its transformed version, which together show the
transformation rule. After that, you will see the query object that requires
the same transformation. Apply the exact same transformation to the query
object and generate the result. Output only the transformed query object as a
single standalone image, matching the visual style and size of the
demonstration result. Do not include the demonstration pair or the original
query in your output.
\end{quote}

\paragraph{Transformation-based / Rule transfer (multi-pair grid items).}
Items in the rule transfer subtask whose rule is demonstrated by multiple
abstract grid-level input-output pairs use a separate instruction:
\begin{quote}\itshape\small
You are solving an abstract visual reasoning puzzle. You will first see one
or more input-output example pairs that demonstrate a transformation rule,
each example shows an input grid followed by its corresponding output grid.
After the examples, you will see a test input grid. Study the examples to
discover the abstract rule, then apply it to the test input. Generate exactly
one output grid for the test input as a standalone image. Do not reproduce
the example pairs or the test input, output only the answer grid, matching
the size, color format, and style of the example output grids.
\end{quote}

\paragraph{Pattern \& Structure / Matrix reasoning.}
\begin{quote}\itshape\small
You are solving a visual matrix puzzle. The input image shows a grid with one
cell missing. Reason over the visual pattern across all rows and columns,
infer what the missing cell must contain, and generate exactly one standalone
image of the missing cell only --- not the full grid. The output must be
cropped to the missing cell. Do not redraw surrounding rows, columns, or the
whole matrix. Match the diagram style of the other cells closely. Do not add
text, labels, option numbers, borders, or extra decorations.
\end{quote}

\paragraph{Pattern \& Structure / Visual-spatial reasoning.}
The visual-spatial reasoning subtask covers a heterogeneous set of activities
(maze solving, path tracing, counting, identification of unique or
duplicate elements, fill-in-the-blank, three-dimensional inspection). Each
activity is associated with a tailored instruction. All variants share a
common output protocol: \emph{the model must output the input image
unchanged except for the indicated annotation} (e.g.\ a red circle, a
red/black line, or a numeric/symbolic answer rendered in black at the
designated blank). The complete list of variants follows.

\begin{description}\itemsep1pt\small

\item[Maze solving.] You are given an image showing a maze with a designated
entrance and a designated exit. Find the correct path through the maze that
connects the entrance to the exit without crossing any walls. Generate an
output image identical to the input, with the correct solution path drawn in
red from the entrance to the exit.

\item[Shortest path on a transit map.] You are given an image showing a
transit network map with stations connected by routes. Two specific stations
are identified as the start and end points. Determine the shortest continuous
route connecting these two stations using the available connections in the
map. Generate an output image identical to the input, with the shortest path
traced in black from the start station to the destination station.

\item[Shortest path on a typed grid.] You are given an image showing a grid
of cells, each containing a letter or symbol, with a designated start cell
and end cell. Find the shortest path from start to end, moving only
horizontally or vertically, passing only through cells that contain the
specific designated element. Output the input image \emph{exactly as-is}; the
only change is drawing the path as a single continuous red polyline from the
centre of the start cell, through the centre of each intermediate qualifying
cell, to the centre of the end cell.

\item[Continuous line tracing.] You are given an image showing a line that
begins at a specific element and follows a continuous path through the image.
Identify the full extent of this continuous line from its starting point to
its end. Generate an output image identical to the input, with the complete
path of the line traced in red from beginning to end.

\item[Match parts to figures.] You are given an image with a three-column
layout: figures with a missing section are placed on both the left and right
sides, while a column of labelled replacement parts runs down the centre.
Each figure is missing a piece; each replacement part fills exactly one
figure. Determine which centre part correctly completes each figure. Output
the input image \emph{exactly as-is}; the only modification is drawing red
lines connecting every figure to the matching replacement part in the centre
column.

\item[Match and connect by computed value.] You are given an image showing
two sets of items: one with values to be computed or measured, and another
with target values or labels. Compute the relevant value for each item in the
first set and identify which item in the second set it corresponds to.
Generate an output image identical to the input, with red lines connecting
each item on one side to its correct match on the other side.

\item[Connect dots to reconstruct a target.] You are given an image showing a
target pattern on one side and separate component elements on the other,
each with a connection dot. Determine how each component must be paired with
the target to correctly reconstruct the target pattern. Generate an output
image identical to the input, with red lines connecting each component's dot
to the corresponding target dot.

\item[Connect 3D figures to top views.] You are given an image showing
three-dimensional solid figures on one side and two-dimensional top-view
projections on the other. Match each 3D figure to the top-view projection
that correctly represents its appearance from directly above. Generate an
output image identical to the input, with red lines connecting each 3D figure
to its correct top-view projection.

\item[Find unpaired element.] You are given an image containing many elements
where every element has exactly one identical pair, except for one element.
Find the single unpaired element. Output the input image \emph{exactly as-is}
with a red circle drawn around the unpaired element.

\item[Find the two identical elements.] You are given an image containing
multiple distinct visual figures or shapes. Among all of them, exactly two
are identical in shape, size, and orientation. Find these two matching
figures. Generate an output image identical to the input, with a red circle
drawn around each of the two identical figures.

\item[Find the identical pair in a scene.] You are given an image containing
many similar-looking figures. Among all figures present, exactly two are
perfectly identical. Locate these two matching figures and circle each in
red.

\item[Find the unique element.] You are given an image containing multiple
visual elements. Exactly one is unique: it differs from all the others
while the remaining elements share a common characteristic. Output the input
image \emph{exactly as-is} with a red circle around the unique element.

\item[Find dots crossed by a line.] You are given an image showing a path
drawn in blue along with dots positioned throughout the image. Identify every
dot that the blue line passes through or touches. Generate an output image
identical to the input, with a red circle drawn around each dot the blue line
intersects.

\item[Find all dark cubes.] You are given an image showing an arrangement of
cubes or blocks. Some cubes are visually dark or heavily shaded; others are
lighter. Identify every dark cube and circle each in red.

\item[Find all matches in a scene.] You are given an image containing a
scene with multiple objects, some of which are matches with a distinctively
coloured tip. Locate every match and circle each in red.

\item[Count cubes in a 3D structure.] You are given an image showing a
three-dimensional structure composed entirely of stacked unit cubes. Count
all individual unit cubes that make up the structure, including any hidden
cubes required to physically support the visible ones. Generate an output
image identical to the input, with the blank or question mark filled in
black with the correct total cube count.

\item[Count and write per-type counts.] You are given an image showing
several reference figure types alongside a larger area containing many
instances. Count how many times each reference figure type appears in the
counting area, excluding example figures. Generate an output image identical
to the input, with each blank filled in black with the correct count.

\item[Count faces of 3D shapes.] You are given an image showing several
three-dimensional geometric shapes, each with a question mark indicating its
face count. Count the number of faces on each shape. Generate an output
image identical to the input, with each question mark replaced in black by
the correct face count, and the formula at the bottom completed using the
four counted values.

\item[Count and write a single total.] You are given an image showing a scene
containing a specific type of object to be counted, along with an empty box
or blank for the answer. Count every instance of the target object visible
in the scene. Generate an output image identical to the input, with the box
filled in black with the correct count.

\item[Compare quantities and fill in operator.] You are given an image
showing two groups of objects on either side of a blank space. Count the
objects in each group and compare the two quantities. Generate an output
image identical to the input, with the blank filled in black using the
correct comparison symbol ($<$,
$>$, or $=$).

\item[Fill colour-coded blanks.] You are given an image showing blank spaces
each outlined in a distinct colour, along with letters, numbers, or symbols
also rendered in corresponding colours. Match each blank to the symbol that
shares its colour. Generate an output image identical to the input, with each
blank filled in black with the correct corresponding symbol.

\item[Fill shapes with indicated colour.] You are given an image where
coloured indicators are connected to geometric shapes by dashed lines, with
each line indicating which colour should fill the corresponding shape. Fill
each shape with its designated colour. Generate an output image identical to
the input, with each shape filled with the colour indicated by its connected
dashed line.

\end{description}

\paragraph{Scenario-based / Scientific process.}
\begin{quote}\itshape\small
You are given a chemical reaction diagram that uses curved-arrow notation to
represent how electrons move during a mechanistic step. Analyse the electron
flow indicated by the arrows to determine the intermediate or product that
results from this specific reaction step. Generate a structural diagram of
the resulting molecule or intermediate, drawn in standard chemical structure
notation consistent with the input diagram. Output only the resulting
chemical structure as a clean, standalone structural diagram.
\end{quote}

\paragraph{Scenario-based / Scene-temporal inference.}
Two instruction variants are used, depending on whether the input frames
must be \emph{predicted forward} or \emph{re-ordered}.

\textbf{Dynamic scene prediction.}
\begin{quote}\itshape\small
You are given a sequence of images showing a dynamic scene, process, or
activity unfolding over time. Study the progression across all images and
identify the underlying pattern, motion, or causal chain. Generate a single
image depicting what would most plausibly occur at the next moment,
continuing naturally from the final state shown. Your output should be
visually consistent with the setting, subjects, and style of the provided
images.
\end{quote}

\textbf{Chronological ordering.}
\begin{quote}\itshape\small
You are given a set of images depicting events or states from the same
process or narrative, presented in shuffled order. Determine the correct
chronological sequence from earliest to latest. Generate a single composite
image that shows all the input images arranged in correct temporal order
from left to right, maintaining the same visual format and relative
proportions as the input images.
\end{quote}

\paragraph{Scenario-based / Cross-domain analogy.}
Four instruction variants are used, one per cross-domain pairing.

\textbf{Animal visual analogy ($A:B :: C:\,?$).}
\begin{quote}\itshape\small
You are given three animal images forming the first three parts of a visual
analogy. The relationship between the first two animals establishes an
analogy rule --- a specific biological, behavioural, ecological, or
categorical connection. Apply the same relational rule to the third animal to
determine what the fourth animal should be. Generate a single image of the
animal that correctly completes the analogy $A:B :: C:\,?$.
\end{quote}

\textbf{Sports sequence continuation.}
\begin{quote}\itshape\small
You are given a sequence of images depicting different sports or physical
activities that follow a visual or categorical pattern. Study the ordering
or grouping logic across the images and identify the rule or pattern that
connects them. Generate a single image of the sport or athletic activity
that should come next in the sequence, matching the visual style and framing
of the provided images.
\end{quote}

\textbf{Species distribution forecasting.}
\begin{quote}\itshape\small
You are given a series of maps or spatial distribution diagrams showing how
a species' range or density has changed across consecutive time periods.
Study the spatial trends and progression visible across all provided maps.
Based on the observed pattern of change, predict what the distribution would
look like in the next time period. Generate a single map image showing the
predicted distribution, using the same visual format and geographic layout
as the input maps.
\end{quote}

\textbf{Plant disease analogy.}
\begin{quote}\itshape\small
You are given images of plant leaves all affected by the same disease or
condition. Study the visual symptoms shared across all provided images ---
such as discolouration patterns, lesion shapes, texture changes, or other
visible signs. Generate a single new image of a plant leaf clearly showing
the same type of disease with visually consistent symptoms. The leaf variety
and setting may differ from the examples, but the disease appearance must
be recognisably the same type.
\end{quote}

\paragraph{Scenario-based / Style preference.}
Two instruction variants are used, depending on the style domain.

\textbf{Product-style continuation.}
\begin{quote}\itshape\small
You are given images of shoes that reflect a consistent style preference or
aesthetic. Study the design language, construction style, materials, and
overall look shared across the provided examples. Generate a single image of
a different pair of shoes that fits the same style profile and would be a
natural addition to this collection. Output a clean, product-style image of
the shoes.
\end{quote}

\textbf{Artist-style continuation.}
\begin{quote}\itshape\small
You are given a set of artworks all created by the same artist. Study the
visual style, technique, composition, colour palette, and subject matter
shared across these works to understand the artist's distinctive approach.
Generate a new artwork that would plausibly have been created by the same
artist, faithfully reflecting their characteristic style. Output a single
painting or illustration that exhibits the same stylistic characteristics.
\end{quote}

\section{Human Validation of Automatic Evaluation}
\label{app:human_validation}

We use two human studies for different purposes. The calibration study
validates automatic scoring on model-generated outputs, while the pilot study
checks whether the benchmark items are well-posed and solvable by humans.

\begin{table}[h]
\centering
\caption{Summary of the two human studies.}
\label{tab:human_study_summary}
\small
\setlength{\tabcolsep}{5pt}
\resizebox{\textwidth}{!}{%
\begin{tabular}{l l c c c l}
\toprule
Study & Purpose & \# Items & \# Annotators & Ratings per item & Output judged \\
\midrule
Human calibration & Validate LLM judge & 500 & 12 & 9 model outputs & Model-generated answers \\
Human pilot & Verify task well-posedness & 66 & 3 & 1 answer per annotator & Human-drawn or web-retrieved answers \\
\bottomrule
\end{tabular}
}
\end{table}

\begin{table}[h]
\centering
\caption{Inter-annotator agreement in the human calibration study.}
\label{tab:human_iaa}
\small
\setlength{\tabcolsep}{8pt}
\begin{tabular}{lcc}
\toprule
Dimension & Krippendorff's $\alpha$ & Ratings within one point \\
\midrule
Composite score & 0.791 & 74.4\% \\
Visual quality & 0.553 & 75.3\% \\
Structural alignment & 0.558 & 61.3\% \\
Reasoning correctness & 0.760 & 82.5\% \\
\bottomrule
\end{tabular}
\end{table}

\begin{table}[h]
\centering
\caption{Subtask-level agreement between Gemini judge scores and independent
human ratings.}
\label{tab:subtask_human_correlation}
\scriptsize
\setlength{\tabcolsep}{4pt}
\begin{tabular}{lcc}
\toprule
Subtask & Pearson composite & Spearman composite \\
\midrule
Overall & 0.766 & 0.711 \\
Interaction concept & 0.903 & 0.849 \\
Abstract concept & 0.691 & 0.680 \\
Geometric analogy & 0.539 & 0.503 \\
Attribute analogy & 0.543 & 0.627 \\
Rule transfer & 0.555 & 0.630 \\
Matrix reasoning & 0.627 & 0.675 \\
Visual-spatial reasoning & 0.733 & 0.580 \\
Scientific process & 0.601 & 0.770 \\
Cross-domain analogy & 0.757 & 0.679 \\
Style preference & 0.519 & 0.581 \\
Scene-temporal inference & 0.555 & 0.644 \\
\bottomrule
\end{tabular}
\end{table}

\section{Additional Diagnostic Analyses}
\label{app:diagnostic}

\subsection{Text Thinking and Self-Reflection}
\label{app:thinking_reflection}

We further evaluate two interventions on the same matched 200-item
subset used in Section~\ref{sec:analysis}: explicit text thinking before
image generation (H) and one or two rounds of image self-reflection
(SR-1/SR-2). All outputs are scored blindly with the paper's image
rubric and reported in the same 0--100 unit as Table~\ref{tab:main_results}.

\begin{table}[h]
\centering
\caption{Effect of explicit text thinking and self-reflection. Each
$\Delta$ is paired against the matched direct-generation baseline from
the corresponding evaluation pass.}
\label{tab:thinking_reflection}
\small
\setlength{\tabcolsep}{5pt}
\resizebox{\textwidth}{!}{%
\begin{tabular}{llcc}
\toprule
Model & Strategy & Judge score & $\Delta$ vs. matched direct baseline \\
\midrule
BAGEL & Direct generation (D) & 33.6 & -- \\
      & Text thinking before generation (H) & 39.7 & +6.1 [+3.3, +9.0] \\
      & One-round self-reflection (SR-1) & 36.0 & +3.7 [+1.3, +6.3] \\
      & Two-round self-reflection (SR-2) & 29.4 & -3.2 [-5.9, -0.6] \\
\midrule
Gemini~3.1~Flash Image Preview & Direct generation (D) & 59.1 & -- \\
      & Text thinking before generation (H) & 66.5 & +7.4 [+2.8, +12.0] \\
      & One-round self-reflection (SR-1) & 64.7 & +7.0 [+2.6, +11.2] \\
      & Two-round self-reflection (SR-2) & 67.3 & +9.4 [+5.4, +13.4] \\
\bottomrule
\end{tabular}
}
\end{table}

The intervention effects are strongly model-dependent. Explicit text
thinking improves BAGEL and Gemini, and also improves the Emu3.5
pipeline in Table~\ref{tab:matched_decomposition}, but it does not
significantly improve the Qwen-Image-Edit pipeline. Self-reflection is
similarly heterogeneous: two rounds improve Gemini but degrade BAGEL,
showing that visual feedback is not uniformly reliable across current
generators.

\subsection{Per-Subtask Cascade Decomposition}
\label{app:cascade}

As a complementary single-model diagnostic, we decompose
Gemini~3.1~Flash Image Preview into perception, reasoning, and
generation stages on a stratified $n{=}100$ subset. Perception asks the
model to describe the input images without solving; reasoning asks it to
commit to a textual answer; generation is the original image output.
Each stage is scored on a 0--5 scale and passes at $\tau{=}3$.

\begin{table}[h]
\centering
\caption{Per-subtask cascade decomposition for Gemini~3.1~Flash Image
Preview. $\mathbb{P}(P)$, $\mathbb{P}(R\mid P)$, and
$\mathbb{P}(G\mid R)$ are pass rates at $\tau{=}3$. The lowest column
per row identifies the bottleneck.}
\label{tab:cascade}
\small
\setlength{\tabcolsep}{6pt}
\resizebox{\textwidth}{!}{%
\begin{tabular}{llcccc}
\toprule
Family & Subtask & $\mathbb{P}(P)$ & $\mathbb{P}(R\mid P)$ &
$\mathbb{P}(G\mid R)$ & Bottleneck \\
\midrule
Concept-based        & interaction concept       & 100\% & 80\% & 88\% & Reasoning  \\
                     & abstract concept          & 100\% & 57\% & 75\% & Reasoning  \\
\midrule
Transformation-based & geometric analogy         &  42\% & 40\% & 60\% & Reasoning \\
                     & attribute analogy         & 100\% & 25\% & 33\% & Reasoning  \\
                     & rule transfer             & 100\% & 17\% &  0\% & \textbf{Generation} \\
\midrule
Pattern \& Structure & matrix reasoning          &  93\% & 54\% & 75\% & Reasoning  \\
                     & visual-spatial reasoning  &  83\% & 80\% & 20\% & \textbf{Generation} \\
\midrule
Scenario-based       & scientific process        & 100\% & 50\% & 33\% & \textbf{Generation} \\
                     & scene-temporal inference  & 100\% & 33\% &  0\% & \textbf{Generation} \\
                     & cross-domain analogy      & 100\% & 33\% & 100\% & Reasoning  \\
                     & style preference          & 100\% & 67\% & 100\% & Reasoning  \\
\bottomrule
\end{tabular}
}
\end{table}

\section{Limitations}
\label{sec:limitaions}

While \benchname~provides a rigorous framework for evaluating Reasoning-driven Image Generation, we identify several avenues for future expansion.

\begin{itemize}
\item \textbf{Dataset Scale vs. Quality:} First, to ensure the highest standards of logical complexity and annotation accuracy, \benchname~currently prioritizes 2,000 hand-curated items over larger-scale, procedurally generated alternatives. While this size is sufficient for statistically significant benchmarking, expanding the volume while maintaining human-level precision remains a continuous objective.

\item \textbf{Automated Evaluation Nuances:} Second, although our LLM-as-a-judge framework (leveraging Gemini-3.1-Flash) demonstrates strong alignment with human experts, automated metrics may occasionally prioritize stylistic coherence alongside strict logical adherence. We encourage researchers to complement these scores with qualitative analysis to capture the full spectrum of reasoning performance.

\item \textbf{Scope of Modality:} Lastly, the current iteration of our benchmark focuses on static image synthesis. As the field evolves toward generative dynamics, extending the benchmark with new components that require maintaining temporal and spatial consistency in video and 3D reasoning is a promising direction for future versions of this work.
\end{itemize}

\section{Broader Impact}
\label{sec:Boarder_Impact}

\benchname is designed as an evaluation benchmark for reasoning-driven image generation. 
Its main positive impact is to provide a diagnostic framework for studying whether multimodal generative models can produce visually correct answers through reasoning, rather than relying only on surface-level visual plausibility. 
This may support future work on more reliable multimodal systems for education, scientific visualization, diagrammatic reasoning, and other settings where generated images should be logically grounded.

The potential negative impacts mainly arise from downstream use of stronger reasoning-driven image generation systems. 
Such systems could be misused to generate misleading visual evidence, persuasive disinformation, or plausible but incorrect diagrams and explanations. Even without malicious intent, visually convincing but logically wrong generations may lead users to overtrust model outputs, especially in high-stakes domains. 
Our benchmark does not itself deploy a generative model, but it may inform the development of stronger systems. 
We therefore recommend that future applications include human oversight, transparent disclosure of generated content, and task-specific validation before use in sensitive settings.

\section{Dataset Provenance}
Table~\ref{tab:subtask_provenance} provides the detailed provenance of each subtask in RIG-Bench. For each subtask, we report its source dataset(s), the original data split from which examples were drawn, the size of the corresponding source pool considered during benchmark construction, and the number of examples included in the final benchmark. The resulting 11 subtasks contain 2,000 examples in total.

\begin{table*}[h]
\centering
\caption{
Subtask-level provenance of RIG-Bench.
}
\label{tab:subtask_provenance}
\resizebox{\textwidth}{!}{
\begin{tabular}{llllrr}
\toprule
\textbf{Family} & \textbf{Subtask} & \textbf{Source dataset(s)}
& \textbf{Original split} & \textbf{Original pool} & \textbf{Final \#} \\
\midrule
Concept-based
& Interaction concept
& Bongard-HOI~\citep{bongard_hoi}
& Train
& 216 & 216 \\

Concept-based
& Abstract concept
& Bongard-RWR+~\citep{bongard_rwr}
& Train
& 152 & 152 \\

\midrule
Transformation-based
& Geometric analogy
& KiVA + KiVA-adults~\citep{kiva}
& Full
& 1,150 & 240 \\

Transformation-based
& Attribute analogy
& KiVA + KiVA-adults~\citep{kiva}
& Full
& 1,000 & 240 \\

Transformation-based
& Rule transfer
& ARC-AGI~\citep{arc,arc_agi_2}
& Train
& 111 & 109 \\

\midrule
Pattern \& Structure
& Matrix reasoning
& MaRs-VQA~\citep{mars_vqa}
& Train
& 1,440 & 326 \\

Pattern \& Structure
& Visual-spatial reasoning
& BabyVision~\citep{babyvision}
& Train
& 131 & 131 \\

\midrule
Scenario-based
& Scientific process
& EMMA~\citep{emma}
& Test
& 105 & 96 \\

Scenario-based
& Scene-temporal inference
& MMR-Life~\citep{mmr_life}
& Test
& 423 & 180 \\

Scenario-based
& Cross-domain analogy
& MMR-Life~\citep{mmr_life}
& Test
& 568 & 180 \\

Scenario-based
& Style preference
& MMR-Life~\citep{mmr_life}
& Test
& 368 & 130 \\

\midrule
\multicolumn{4}{r}{\textbf{Total final benchmark}} & & \textbf{2,000} \\
\bottomrule
\end{tabular}
}
\end{table*}

\section{User Study Interface}
\label{sec:user_study}

\begin{figure*}[h]
  \centering
  \includegraphics[width=\linewidth]{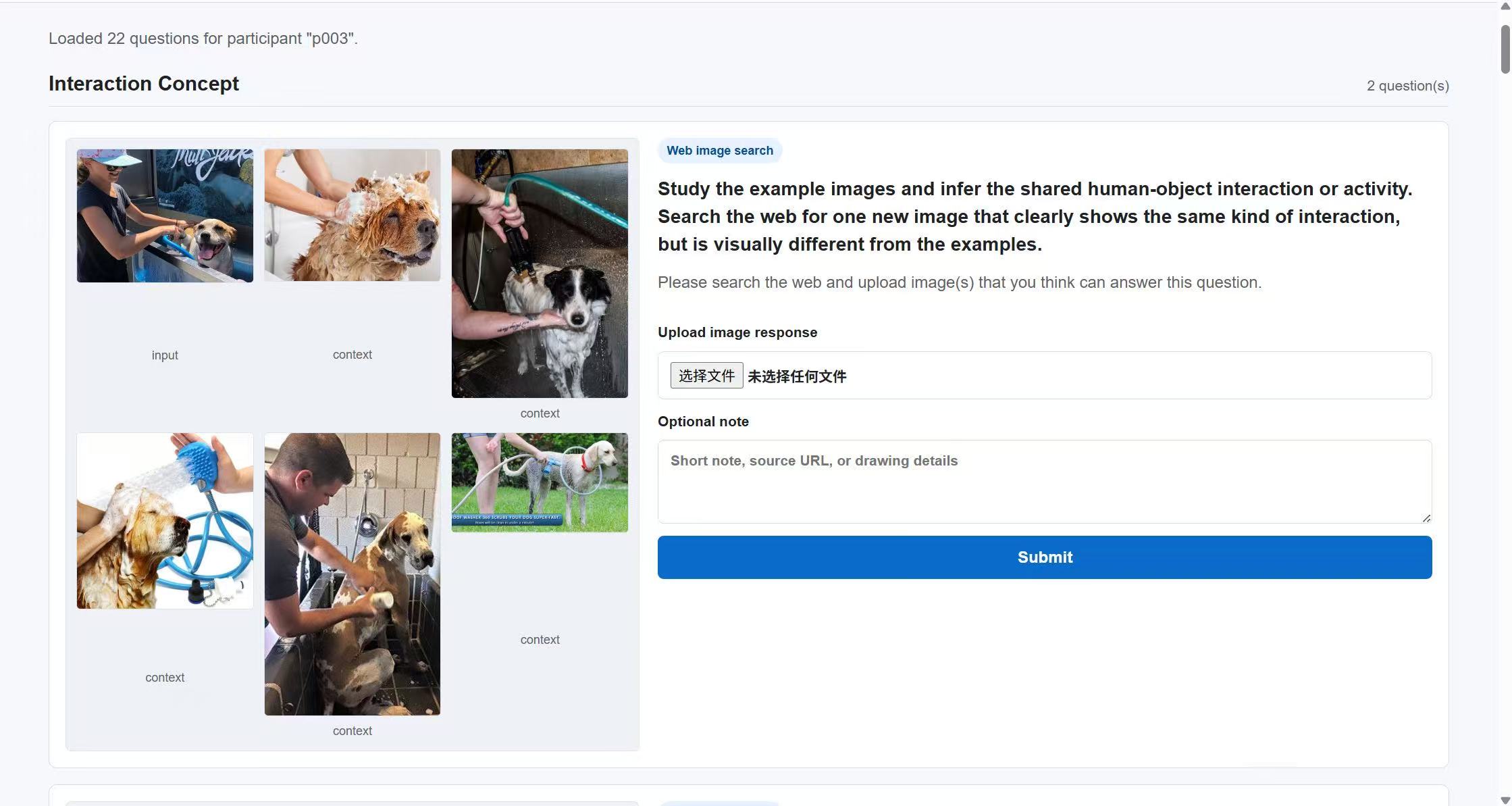}
  \includegraphics[width=\linewidth]{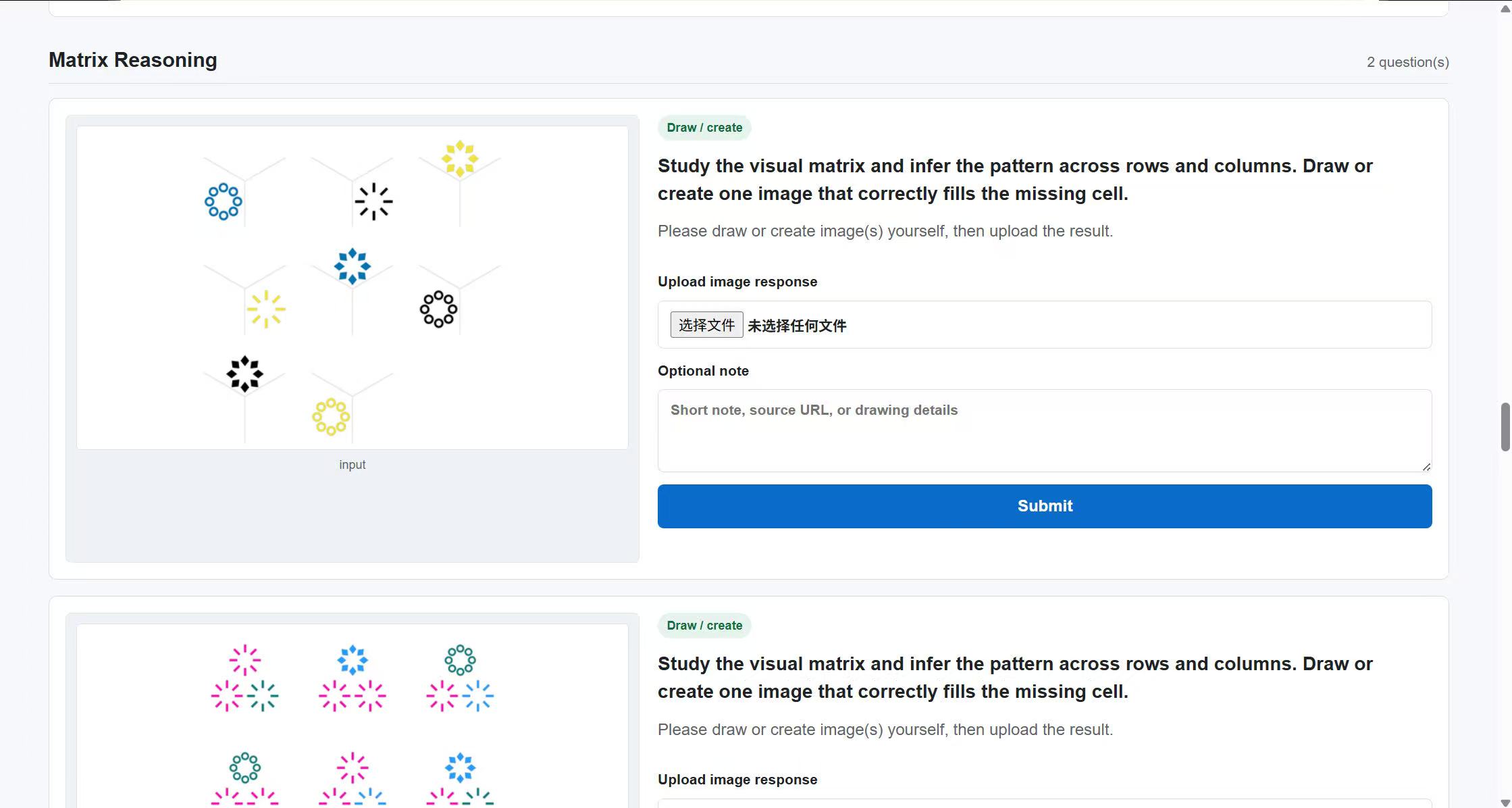}
  \caption{Screenshots of the user study interface.}
  \label{fig:user_study_cite}
\end{figure*}

Figure~\ref{fig:user_study_cite} shows screenshots of the interface used in our formative human pilot. 
The instruction shown to participants follows the same task prompt format described in the main benchmark setup, where users are asked to infer the correct target image from the given visual context. They are asked to either draw it or retrieve a matching image from the web.

\section{Additional Qualitative Examples}
\label{app:qualitative_examples}

\begin{figure*}[h]
  \centering
  \includegraphics[width=\linewidth]{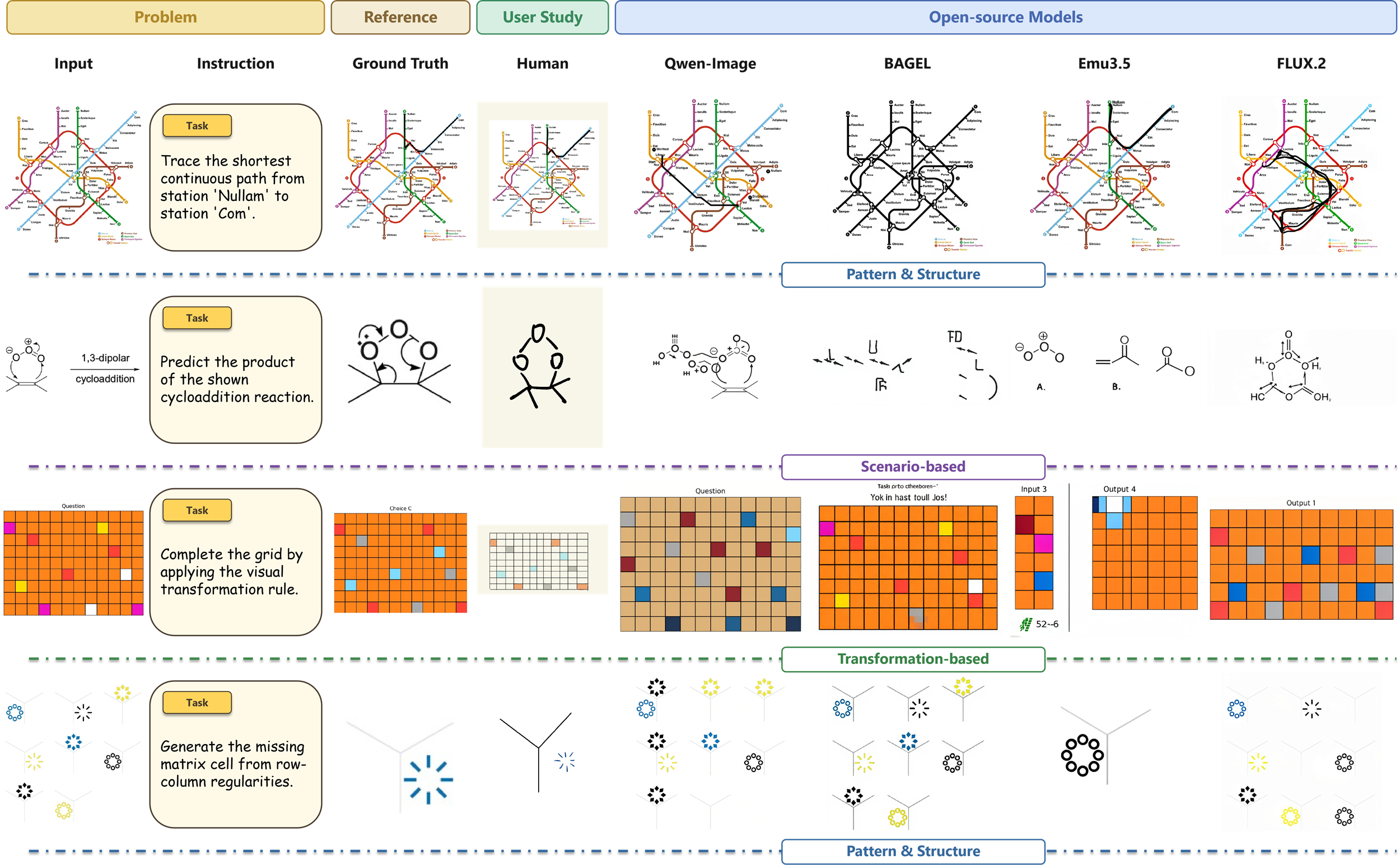}
  \includegraphics[width=\linewidth]{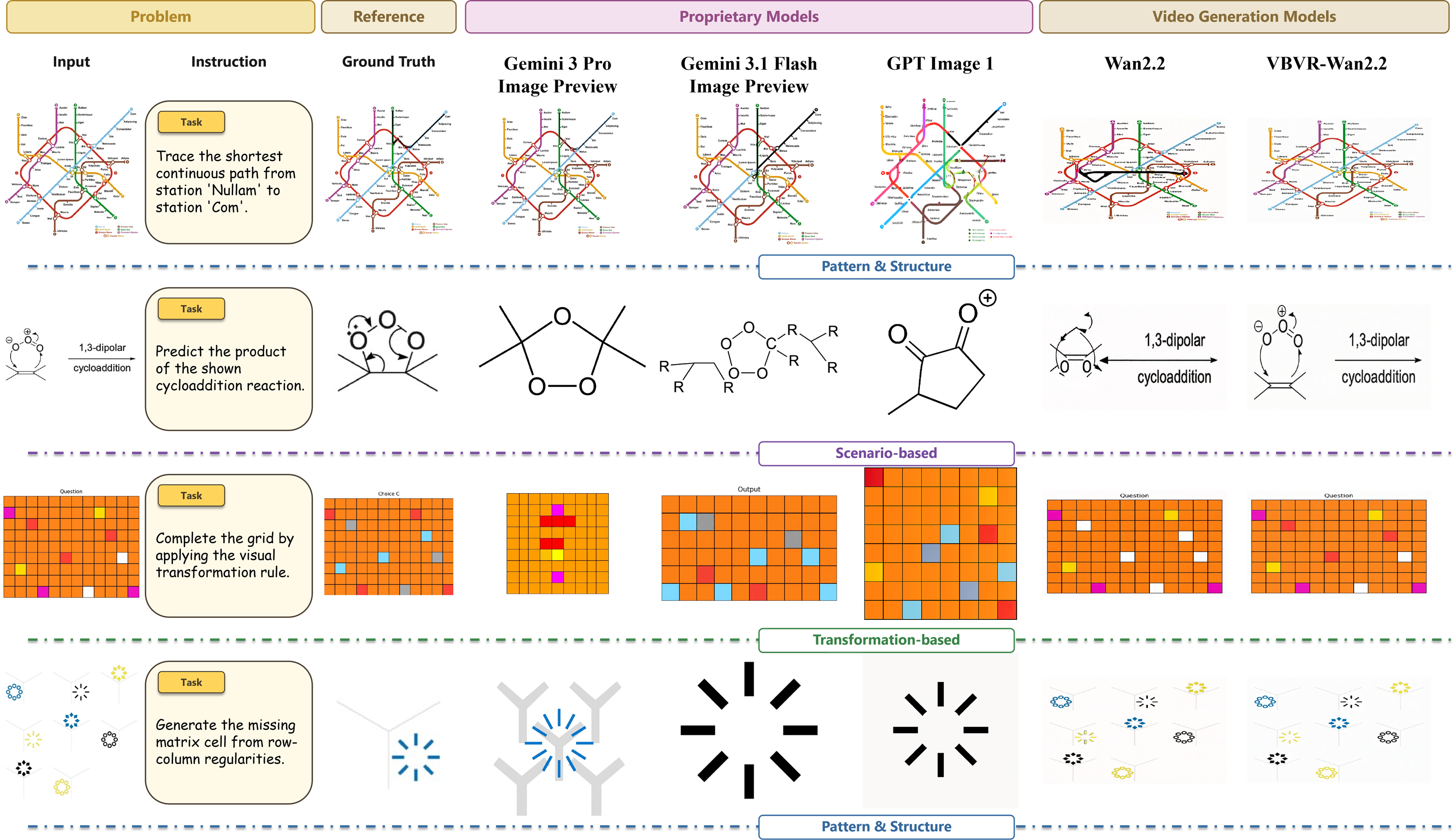}
  \caption{Extended qualitative example group 1.}
  \label{fig:appendix_example_group1}
\end{figure*}

\begin{figure*}[h]
  \centering
  \includegraphics[width=\linewidth]{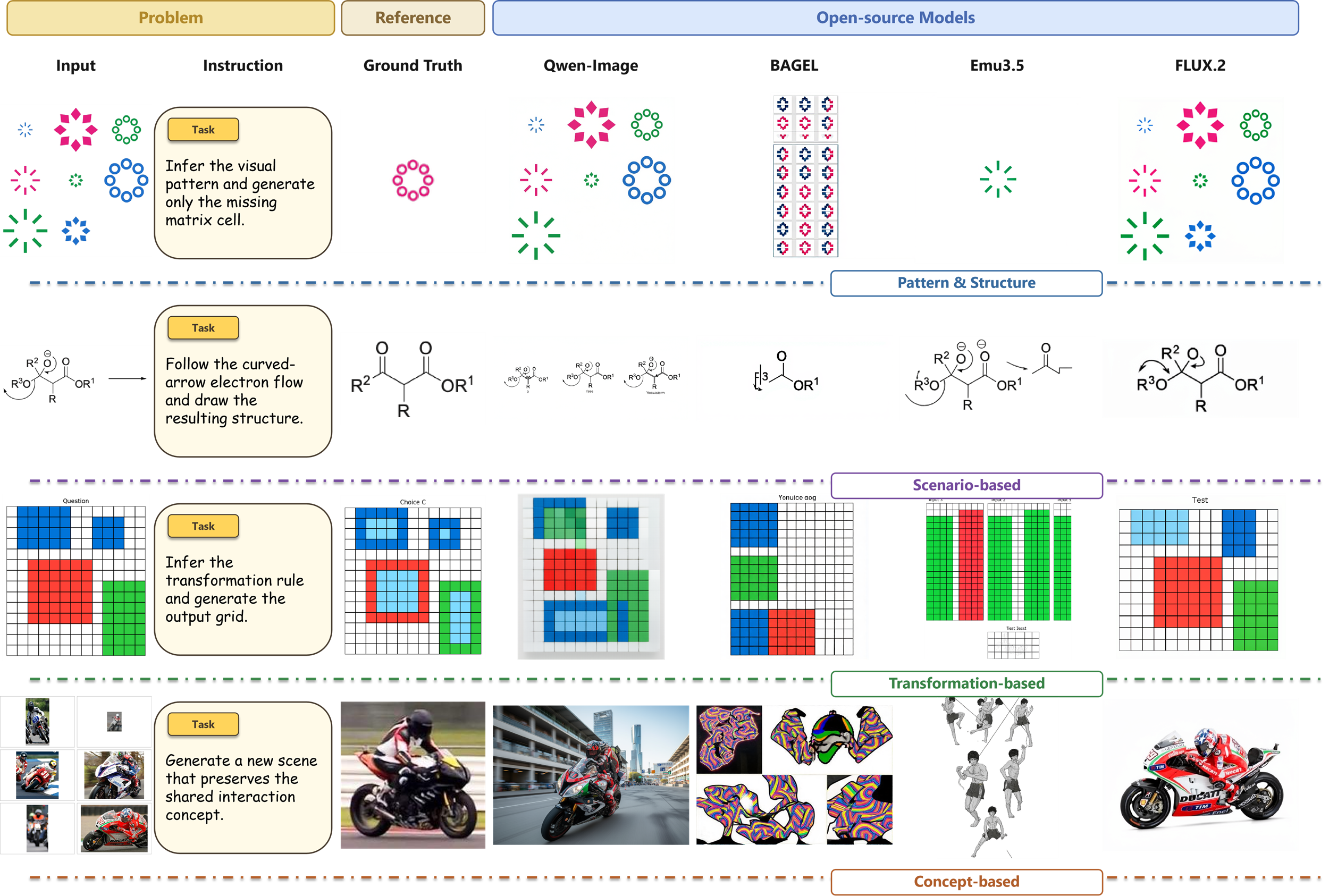}
  \includegraphics[width=\linewidth]{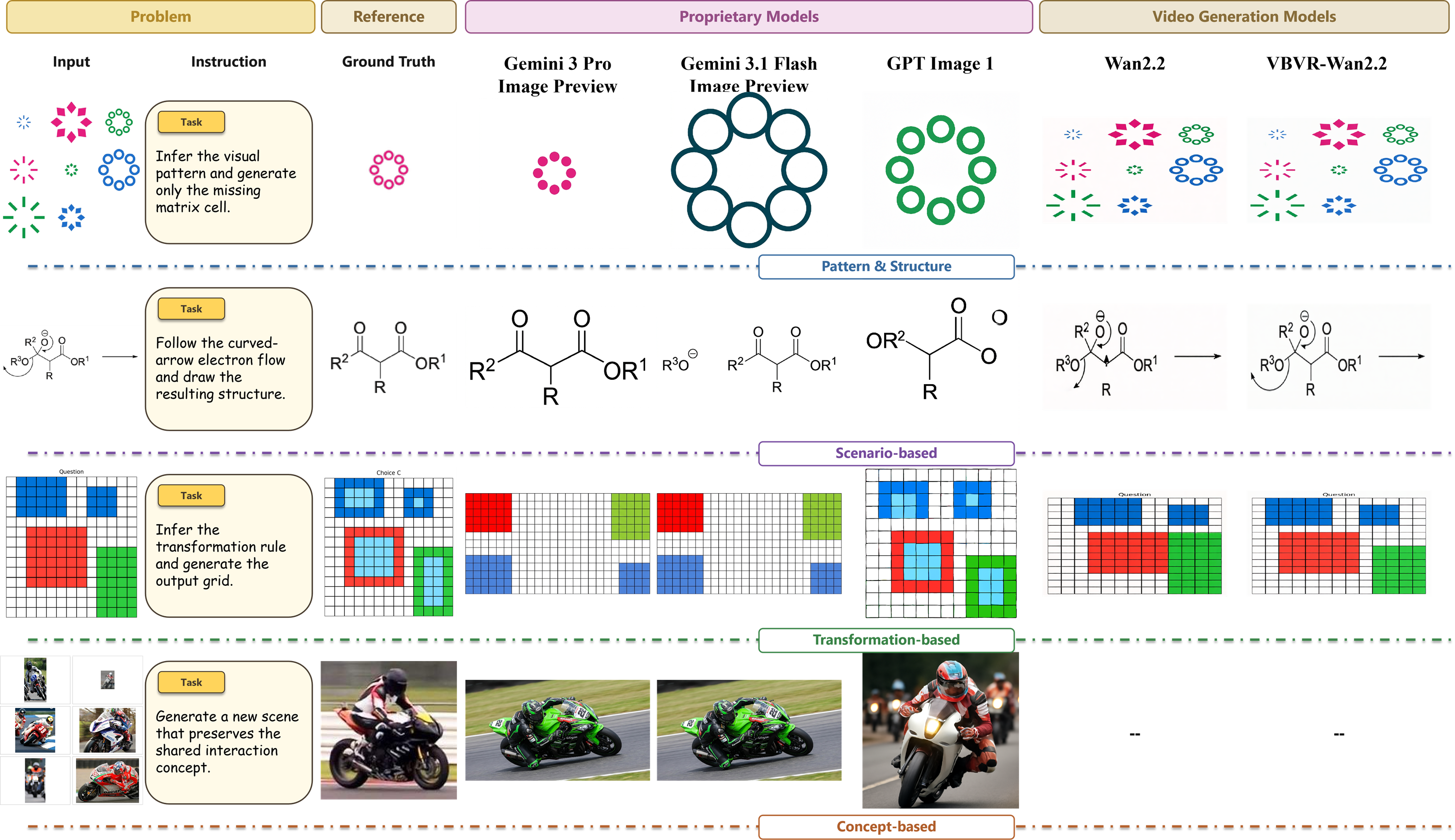}
  \caption{Extended qualitative example group 2.}
  \label{fig:appendix_example_group2}
\end{figure*}

\begin{figure*}[h]
  \centering
  \includegraphics[width=\linewidth]{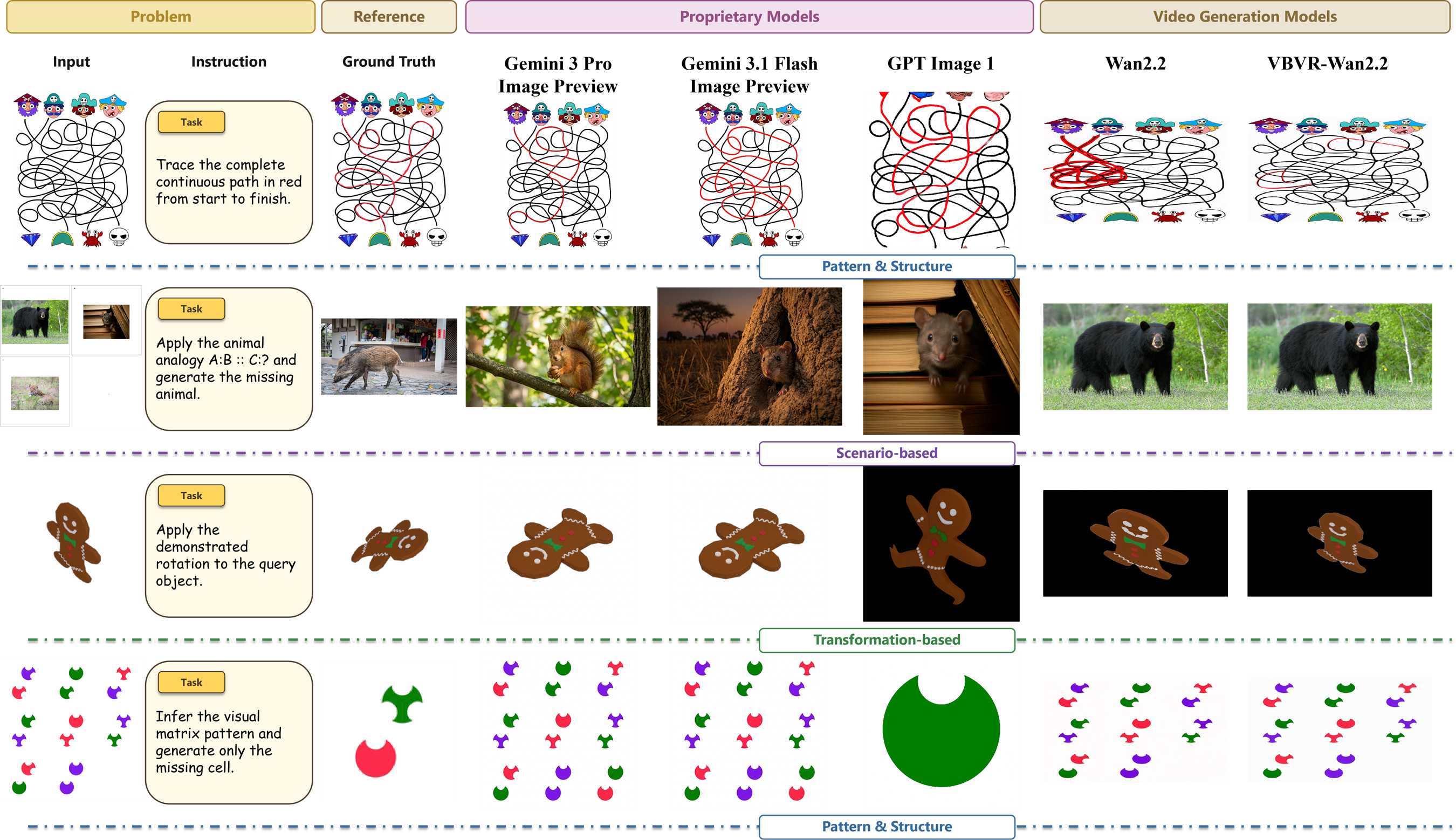}
  \includegraphics[width=\linewidth]{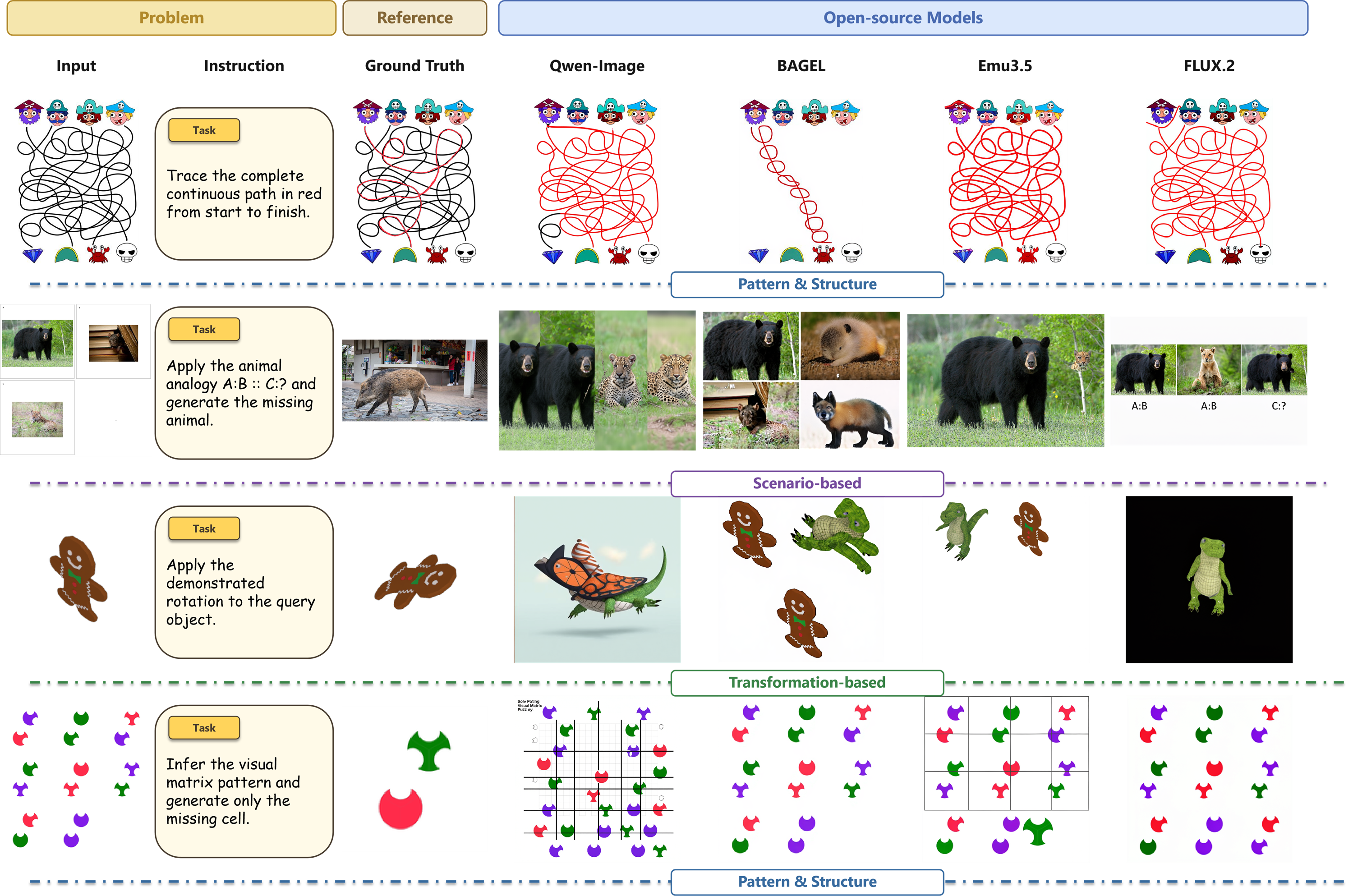}
  \caption{Extended qualitative example group 3.}
  \label{fig:appendix_example_group3}
\end{figure*}

Figures~\ref{fig:appendix_example_group1},
\ref{fig:appendix_example_group2}, and
\ref{fig:appendix_example_group3} provide additional qualitative examples demonstrating representative successes and failure modes of \benchname.



\clearpage

\end{document}